\documentclass[9.5pt,journal,compsoc]{IEEEtran}

\ifCLASSOPTIONcompsoc
  \usepackage[nocompress]{cite}
\else
  \usepackage{cite}
\fi

\ifCLASSINFOpdf
\else
\fi

\usepackage{newpxtext,newpxmath}

\usepackage{algpseudocode}
\usepackage{algorithm}
\usepackage{amsmath,amssymb,amsthm,amsfonts}
\usepackage{breqn}
\usepackage{mathtools}
\usepackage{cases}
\usepackage{mathrsfs}
\usepackage{accents}
\usepackage{epstopdf}
\usepackage{color}
\usepackage{tabu}
\usepackage{stfloats}
\usepackage{mathptmx}
\usepackage{epstopdf}

\usepackage{caption}
\usepackage{graphicx}
\usepackage{times}
\usepackage{array}
\usepackage{subfig}
\usepackage[T1]{fontenc}
\usepackage[latin1]{inputenc}
\usepackage{url}
\usepackage[normalem]{ulem}
\useunder{\uline}{\ul}{}
\usepackage{balance}
\usepackage{rotating}
\usepackage{scrextend}
\usepackage{multirow}

\newlength{\dhatheight}

\newcommand{\etal}{\textit{et al.}}
\newcommand{\Fig}{Fig.}

\newcolumntype{P}[1]{>{\centering\arraybackslash}p{#1}}
\newcolumntype{M}[1]{>{\centering\arraybackslash}m{#1}}
\makeatletter
\newcommand*\bigcdot{\mathpalette\bigcdot@{1}}
\newcommand*\bigcdot@[2]{\mathbin{\vcenter{\hbox{\scalebox{#2}{$\m@th#1\bullet$}}}}}
\makeatother

\makeatletter
\renewcommand{\ALG@beginalgorithmic}{\small}
\makeatother

\begin{document}
\title{ConiVAT: Cluster Tendency Assessment and Clustering with Partial Background Knowledge}



\author{Punit~Rathore,~James~C.~Bezdek,~Paolo~Santi~and~Carlo~Ratti
\IEEEcompsocitemizethanks{\IEEEcompsocthanksitem Punit Rathore, Paolo Santi, and Carlo Ratti are with the Senseable City Lab at Massachusetts Institute of Technology, Cambridge, MA, 02139, USA. Paolo Santi is also affiliated with the Institute of Informatics and Telematics of CNR, Pisa, Italy.
\protect\\
E-mails:  \{prathore,psanti,ratti\}@mit.edu
\IEEEcompsocthanksitem James C. Bezdek is with the School of Computing and Information
Systems, The University of Melbourne, Victoria, Australia.
\protect
E-mail: jcbezdek@gmail.com.
}
}

\markboth{IEEE TRANSACTIONS ON KNOWLEDGE AND DATA ENGINEERING}
{Rathore \MakeLowercase{\textit{et al.}}: ConiVAT: Cluster Tendency Assessment and Clustering with Partial Background Knowledge}

\IEEEtitleabstractindextext{%
\begin{abstract}
The VAT method is a visual technique for determining the potential cluster structure and the possible number of clusters in numerical data. Its improved version, iVAT, uses a path-based distance transform to improve the effectiveness of VAT for "tough" cases. 
 Both VAT and iVAT have also been used in conjunction with a \textit{single-linkage} (SL) hierarchical clustering algorithm. 
However, they are sensitive to noise and bridge points between clusters in the dataset, and consequently, the corresponding VAT/iVAT images are often in-conclusive for such cases. In this paper, we propose a constraint-based version of iVAT, which we call ConiVAT, that makes use of background knowledge in the form of constraints, to improve VAT/iVAT for challenging and complex datasets. ConiVAT uses the input constraints to learn the underlying similarity metric and builds a minimum transitive dissimilarity matrix, before applying VAT to it. We demonstrate ConiVAT approach to visual assessment and single linkage clustering on nine datasets to show that, it improves the quality of iVAT images for complex datasets, and it also overcomes the limitation of SL clustering with VAT/iVAT due to "noisy" bridges between clusters. Extensive experiment results on nine datasets suggest that ConiVAT outperforms the other three semi-supervised clustering algorithms in terms of improved clustering accuracy.
\end{abstract} 

}

\maketitle

\IEEEdisplaynontitleabstractindextext

\IEEEpeerreviewmaketitle

\section{Introduction}\label{sec:introduction}
Everyday, large amounts of data are generated in many real-life applications from various sources such as IoT networks, smartphones, and social network activities. Cluster analysis~\cite{aggarwal2013data,bezdek2017primer} is a popular approach to extract relevant useful knowledge and internal relationships in such data. In clustering,  datapoints are partitioned in subsets of similar objects according to some notion of similarity. Traditional clustering algorithms are generally applied in an unsupervised fashion where the algorithm has access only to the feature vectors describing each object or dissimilarity matrix between objects~\cite{wagstaff2001constrained}; it does not rely on any background knowledge or other information e.g. labels.  

However, in many situations, some prior background knowledge about the domain or dataset are available, and could be useful in clustering the data~\cite{bair2013semi}. For example, labels of some observations may be known or certain observations may be known to belong to the same or different clusters. The labeled data and unlabeled data often exist together. Consider an email classification problem~\cite{bair2013semi} where a large database of emails is available and only a small subset of which have already been classified as "spam" or "not spam". One may wish to characterize the properties of "spam" emails and identify "spam" emails in the large dataset. Another  example is cancer diagnosis~\cite{qin2019research}, where one may wish to identify genetic clusters that can be used to determine the prognosis of cancer patients. Such clusters would only be of interest if they were associated with the outcome of interest, namely patient survival. To make cancer diagnosis more accurate, some known cancer diagnosis information, including expert opinion guidance, and related cancer conditions may be useful. 

To make use of the background knowledge with unlabeled data, several semi-supervised learning techniques have been proposed. Semi-supervised learning techniques include both semi-supervised classification and semi-supervised clustering~\cite{basu2004probabilistic}.  Though semi-supervised classification techniques make use of partially labeled data, they only work well when partially labeled data is significant and represent all the relevant classes. In contrast, semi-supervised clustering~\cite{bair2013semi,qin2019research} approaches can partition the data using the classes in the initial labeled data, as well as extend and modify the existing set of classes as needed to reflect other patterns in the data~\cite{basu2002semi}. 

Semi-supervised clustering approaches consider the partial background knowledge in the form of constraints such as pairwise constraints~\cite{wagstaff2000clustering,klein2002instance} e.g., two instances must be in the same cluster (\textit{must-link} (ML) constraints) or two instances cannot be in the same cluster (\textit{cannot-link} (CL) constraints) or ordering constraints~\cite{zhao2010hierarchical,zheng2011semi}  e.g., two instances must merge before they merge with another instance.  
  Semi-supervised clustering approaches have been successful in recent years to solve practical problems in many applications, including road detection, image classification, bioinformatics, information retrieval, and speech recognition~\cite{qin2019research}.

Though there exist several semi-supervised clustering approaches~\cite{bair2013semi,qin2019research} in the literature, only a few research efforts~\cite{klein2002instance,zhao2010hierarchical,zheng2011semi,reddy2016semi} have been devoted to hierarchical semi-supervised clustering methods. Unlike partitioning-based semi-supervised clustering methods~\cite{basu2002semi,basu2004probabilistic,wagstaff2001constrained,wagstaff2000clustering,bilenko2004integrating,gao2006semi} in which objective functions can easily be modified to incorporate the constraints, hierarchical clustering methods have no such global objective functions. Moreover, pairwise constraints are not suitable for hierarchical clustering as objects are linked over different hierarchy levels~\cite{zheng2011semi}.


This article focuses on designing a semi-supervised approach for the \textit{visual assessment of clustering tendency} (VAT)~\cite{bezdek2002vat} algorithm. The VAT and improved VAT (iVAT)~\cite{wang2010ivat,havens2012efficient} algorithms produce an image of a reordered dissimilarity matrix that can be used for visual assessment of cluster structure (including the potential number of clusters) in the data. Moreover, VAT reordering is directly related to the clusters produced by \textit{single linkage} (SL)~\cite{gower1969minimum} hierarchical clustering, thus making it feasible to directly extract SL-aligned partitions from VAT/iVAT images~\cite{havens2009vat}. However, like SL, VAT and iVAT images and subsequent clustering are significantly deteriorated by the presence of noise bridges between clusters~\cite{xu2005survey}. Moreover, VAT and iVAT can often be inconclusive, especially if the cluster structure in the data set is complex~\cite{wang2010ivat,kumar2020visual}.

To overcome the above drawbacks of existing semi-supervised hierarchical clustering approaches, we present a semi-supervised approach of iVAT algorithm, that provides both the potential cluster structure and SL-aligned clustering partitions, without needing $k$ as an external input. Specifically, our major contribution in this article are as follows: 
\begin{itemize}
    \item  We propose a constraint-based, semi-supervised approach for iVAT algorithm, which we call ConiVAT, that incorporates metric learning~\cite{xing2003distance} and constraints from partial background knowledge with iVAT in a principled manner. The ConiVAT model seeks an approximate distance metric~\cite{xing2003distance} that satisfies input constraints, using gradient ascent and an iterative projection-based metric learning approach.
    \item We demonstrate the performance of ConiVAT on nine datasets to show that the ConiVAT  overcomes the "noisy bridge" problem of the VAT/iVAT algorithms and also improves the clustering accuracy of SL. Our method also enhances the iVAT image quality to visually present the potential cluster structure and the reliable number of clusters, even for complex datasets. 
    \item  We will compare the clustering performance of ConiVAT with five other semi-supervised hierarchical clustering approaches. 
    \item We will also examine the benefits of each step of the ConiVAT algorithm. Finally, we will study the effect of varying number of constraints in the ConiVAT framework. 
\end{itemize}


The remainder of this paper is organized as follows: Section~\ref{sec:relatedwork} summarizes the literature on semi-supervised clustering approaches. Section~\ref{sec:VATandiVAT} presents a summary of the VAT and iVAT algorithms and their limitations. The proposed algorithm ConiVAT is discussed in Section~\ref{sec:proposed}. Section~\ref{sec:experiments} presents the experiments and results followed by future work and conclusions in Sections~\ref{sec:futurework} and~\ref{sec:conclusion}.


\section{Related Work}\label{sec:relatedwork}
Although the notion of using background knowledge to improve clustering was first brought by~\cite{pedrycz1985algorithms,thompson1992integrating,talavera1999integrating},  
 the concept of constraint-based semi-supervised clustering was introduced in~\cite{wagstaff2000clustering,wagstaff2001constrained}. Wegstaff~\etal~\cite{wagstaff2001constrained} proposed a constraint-based $k$-means algorithm, "COP k-means", by enforcing the constraints during the cluster assignments. Basu~\etal~\cite{basu2002semi} developed a semi-supervised $k$-means clustering algorithm, "constraint $k$-means", by initializing the cluster centres using the partially labeled data and assigning the labeled observations into their known cluster during cluster assignment step, even if they are closer to the mean of other clusters. In~\cite{basu2002semi}, Basu~\etal  recommended an alternative algorithm, called "seeded $k$-means" which is identical to "constraint $k$-means" except  for the hard assignment of labeled data in cluster assignment step.

One drawback of the above algorithms is that they require no constraints to be violated. However, in some situations, violation of some of the constraints may be justifiable if there is strong evidence that the default constraints are incorrect. Basu~\etal~\cite{basu2004active} proposed a pair-wise constraint-based $k$-means algorithm, called "PCK-means", that integrates the must-link and cannot-link constraints into the objective function, and then, seeks to minimize the modified objective function such that most constraints are satisfied. In~\cite{basu2004active}, Basu~\etal also proposed a variant of PCK-means, called "active PCK-means", that selects subsets of the observations to generate robust and informative constraints to maximize the clustering accuracy. Bilenko~\etal~\cite{bilenko2004integrating} proposed an "MPCK-means" clustering algorithm that integrates metric-learning and constraints in the PCK-means framework. 

The majority of the existing semi-supervised clustering algorithms are based on $k$-means or some other forms of partitional clustering methods. Only a few hierarchical semi-supervised clustering algorithms have been proposed in the literature. This is probably because problems must be formulated differently for hierarchical clustering. For example, Kelin~\etal~\cite{klein2002instance} proposed the \textit{constraint complete-linkage} (CCL) algorithm that first imposes the instances level constraints in an input dissimilarity matrix $D$ by lowering the distance between two "must-link" points to $0$ and setting the "cannot-link" pair entries to $max(D)+1$. Then, it propagates "must-link" constraints by computing all-pairs-shortest-path distances. Finally,  \textit{complete linkage} (CL)-based hierarchical clustering is applied to the resulting dissimilarity matrix to obtain the hierarchy.

Miyamoto~\etal~\cite{miyamoto2010semi} require the instances linked by "must-link" constraints to be merged together at the lowest possible level of the hierarchy and the instances separated by "cannot-link" constraints to not be a part of the hierarchy.    Bade~\etal~\cite{bade2006personalized} and Zhao~\etal~\cite{zhao2010hierarchical} use ordering constraints to set the merging order in a hierarchical agglomerative clustering algorithm.

Zheng~\etal~\cite{zheng2011semi} introduced the \textit{must-link-before} (MLB) constraints and proposed two semi-supervised hierarchical clustering techniques based on the ultra-metric dendrogram. The first technique, called "IPotim", learns an ultra-metric distance matrix from the input constraints using an iterative projection-based method, and the second technique, called "UltraTran" integrates the input constraints with the input dissimilarity matrix to obtain an ultra-metric distance matrix using a modified Floyd-Warshall algorithm. Finally, in both the techniques, the CL-based hierarchical clustering is applied to the ultra-metric matrix to obtain the hierarchy. Similar to the CCL, Reddy~\etal~\cite{reddy2016semi} proposed a single-linkage (SL)- based semi-supervised hierarchical clustering approach, called SSL. Like most traditional clustering algorithms, existing semi-supervised algorithms require the number of clusters ($k$) as an input. In this paper, we present the ConiVAT algorithm that provides both the potential cluster structure and SL-aligned clustering partitions, without needing $k$ as an external input. Since the ConiVAT algorithm is a constraint-based version of VAT and iVAT, we briefly discuss them in the next section.
\section{VAT and iVAT and their limitations}\label{sec:VATandiVAT}
Consider a set of $N$ objects $O=\{o_{1},o_{2},...,o_{N} \}$ where each object is represented by a $p$-dimensional feature vector, $\textbf{x}_{i} \in \mathbb{R}^{p}$ in a set of $X = \{\textbf{x}_{1},..,\textbf{x}_{N}\} \in \mathbb{R}^{p}$. Another way to represent the objects in $O$ is a square $N \times N$ dissimilarity matrix $D = [d_{ij}]$, where $d_{ij}$  represents dissimilarity between $o_{i}$ and $o_{j}$, computed using a chosen 'distance metric'.

The VAT algorithm~\cite{bezdek2002vat} reorders the dissimilarity matrix $D$ to $D^{*}$ using a modified Prim's algorithm that finds the minimum spanning tree of a weighted undirected graph. Each pixel of the VAT image $I(D^{*})$, also called \textit{reordered dissimilarity image} (RDI) or cluster heat map, reflects the dissimilarity value between corresponding row and column objects. In a grayscale RDI image, $I(D^{*})$, white pixels represent high dissimilarity, while black represents low dissimilarity. Each object is exactly similar to itself, which results in zero-valued (black) diagonal elements, and non-zero valued off-diagonal elements in $I(D^{*})$. A dark block along the diagonal of RDI is a sub-matrix of "similar" dissimilarity values; therefore, when dark blocks appear along the diagonal of the RDI $I(D^{*})$,  they potentially represent different (ideally, $k$) clusters of objects that are relatively similar to each other.

\begin{figure}
\captionsetup[subfigure]{justification=centering}
\centering
\subfloat[Synthetic data  $N=5000$]{\includegraphics[width=0.16\textwidth]{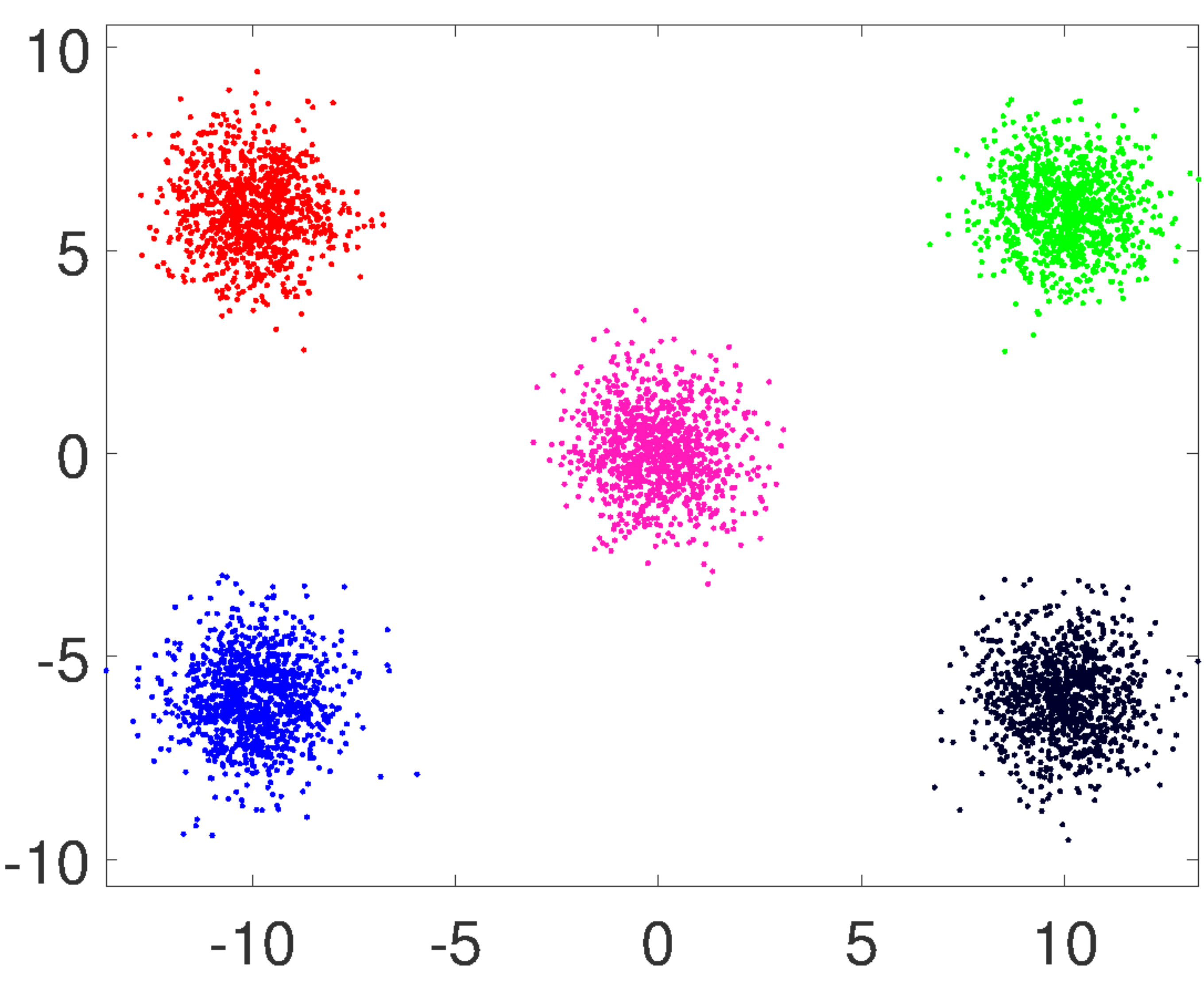}} \hfill
\subfloat[VAT for $N=5000$]{\includegraphics[width=0.16\textwidth]{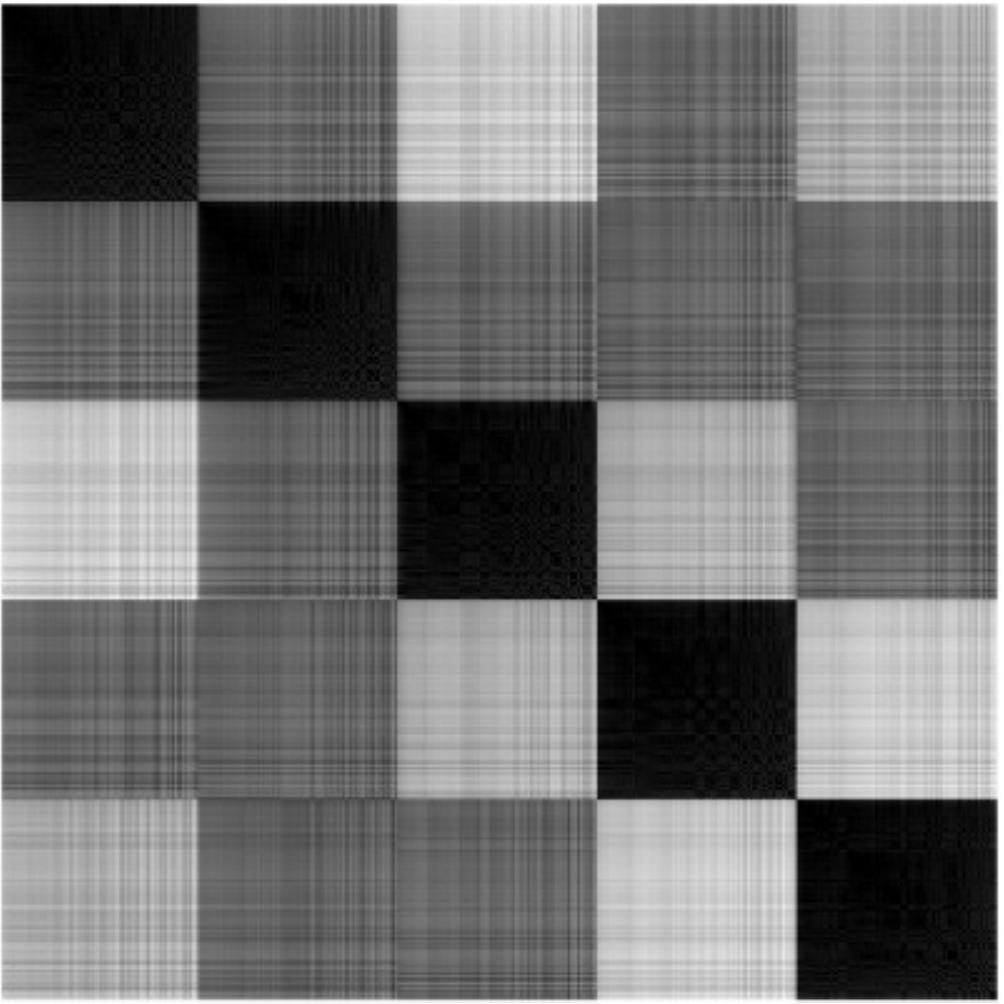}} \hfill
\subfloat[iVAT for $N=5000$]{\includegraphics[width=0.16\textwidth]{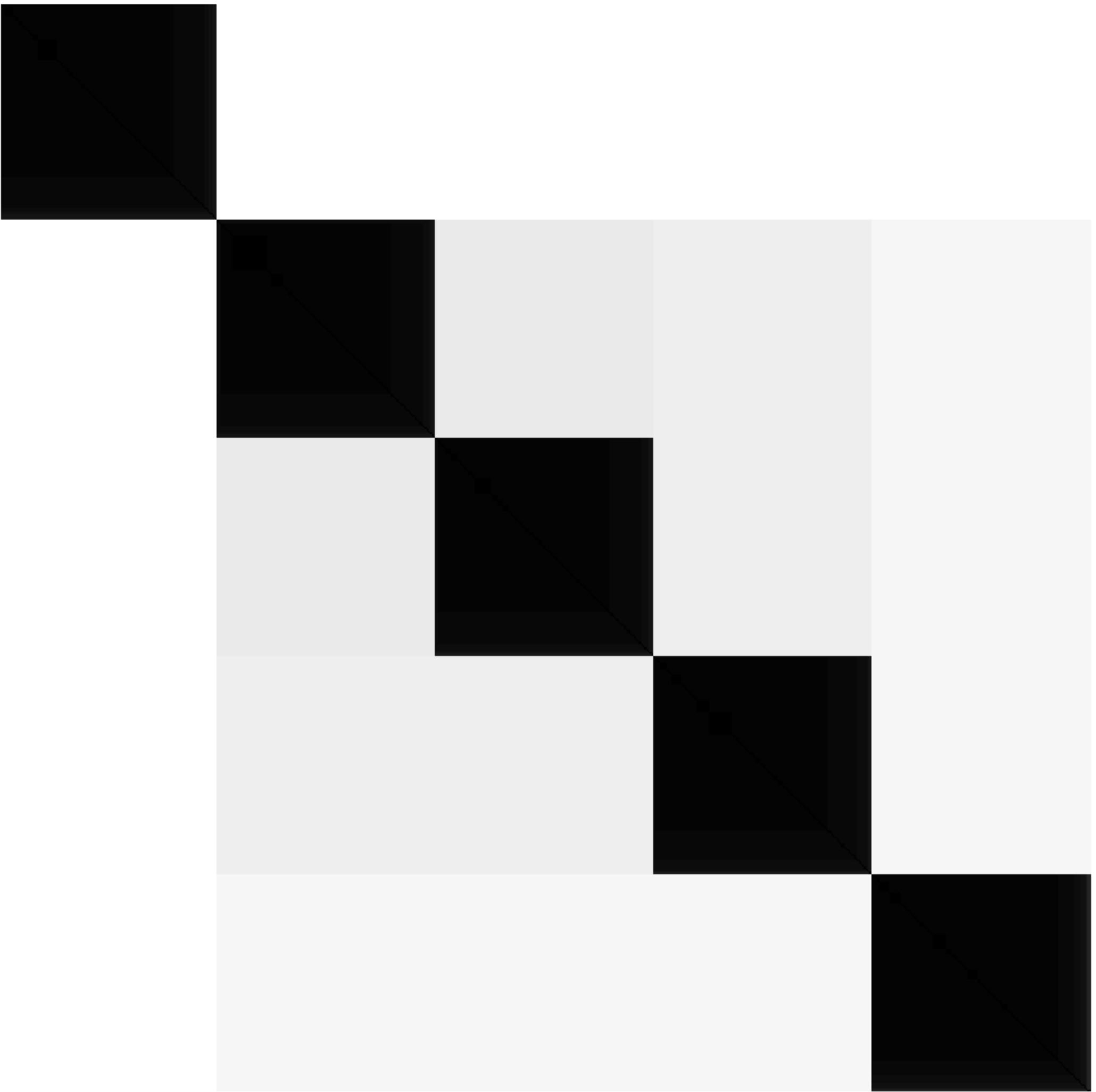}} 
\caption{Data scatterplot, VAT, and iVAT images for a 2D synthetic dataset.}
\label{Fig:VATiVATsiVATImage}
\end{figure}

An algorithm called \textit{improved VAT} (iVAT)~\cite{wang2010ivat,havens2012efficient}  provides a much sharper RDI by replacing input distance $d_{ij}$ in distance matrix $D_N$ by a path-based minimax distance~\cite{fischer2001path}.~\Fig~\ref{Fig:VATiVATsiVATImage} illustrates VAT and iVAT for a 2D synthetic dataset ($N=5000$). While both VAT  and iVAT RDIs show five dark blocks along the diagonal corresponding to the five clusters in the dataset, dark blocks in the iVAT image are much clearer than in the VAT image. The pseudocodes of VAT and iVAT algorithms are well documented in~\cite{rathore2018big,kumar2020visual}, so they will not be replicated here.

Since SL clusters are always diagonally aligned in VAT/iVAT ordered images~\cite{havens2009vat},  $k$-aligned clusters can be obtained with a backpass through the MST by cutting the largest $(k-1)$ edges in the MST. Both VAT and iVAT have following limitations: 
\begin{enumerate}
\item VAT and iVAT RDIs can often be inconclusive,
especially if the cluster structure in the data set is complex.
\item VAT RDI quality is significantly deteriorated by the presence of noise (especially inliers: bridge points between clusters). This shortcoming of VAT is inherited from the SL algorithm (chaining effect), which is the backbone of VAT reordering. 
\end{enumerate}

\section{ConiVAT}\label{sec:proposed}
Given a set of feature vectors  $X = \{\textbf{x}_{1},..,\textbf{x}_{N}\} \in \mathbb{R}^{p}$, and partial background knowledge in the form of a set of "similar" constraints, $\mathbb{S}$ i.e., $ \{( \mathbf{x}_{i},\mathbf{x}_{j}) \} \in \mathbb{S}$, and "dissimilar" constraints, $\mathbb{D}$ i.e., $ \{( \mathbf{x}_{j},\mathbf{x}_{k}) \} \in \mathbb{D}$, the ConiVAT algorithm aims to provide a more reliable RDI and better clustering performance compared to the VAT/iVAT models and existing hierarchical semi-supervised algorithms.

The ConiVAT algorithm aids iVAT by incorporating labeled data in the following three ways: 
\begin{itemize}
    \item Robust constraints generation from partially labeled data when partial information is not in the form of constraints.
    \item Integration of constraints with metric learning to transform the points to a new space in which constraints are satisfied.
    \item Minimum transitive dissimilarity matrix computation using a path-based distance measure to improve cluster structure in the dissimilarity matrix.
\end{itemize}
 
 \begin{figure}
\captionsetup[subfigure]{justification=centering}
\centering
\subfloat[]{\includegraphics[width=0.16\textwidth]{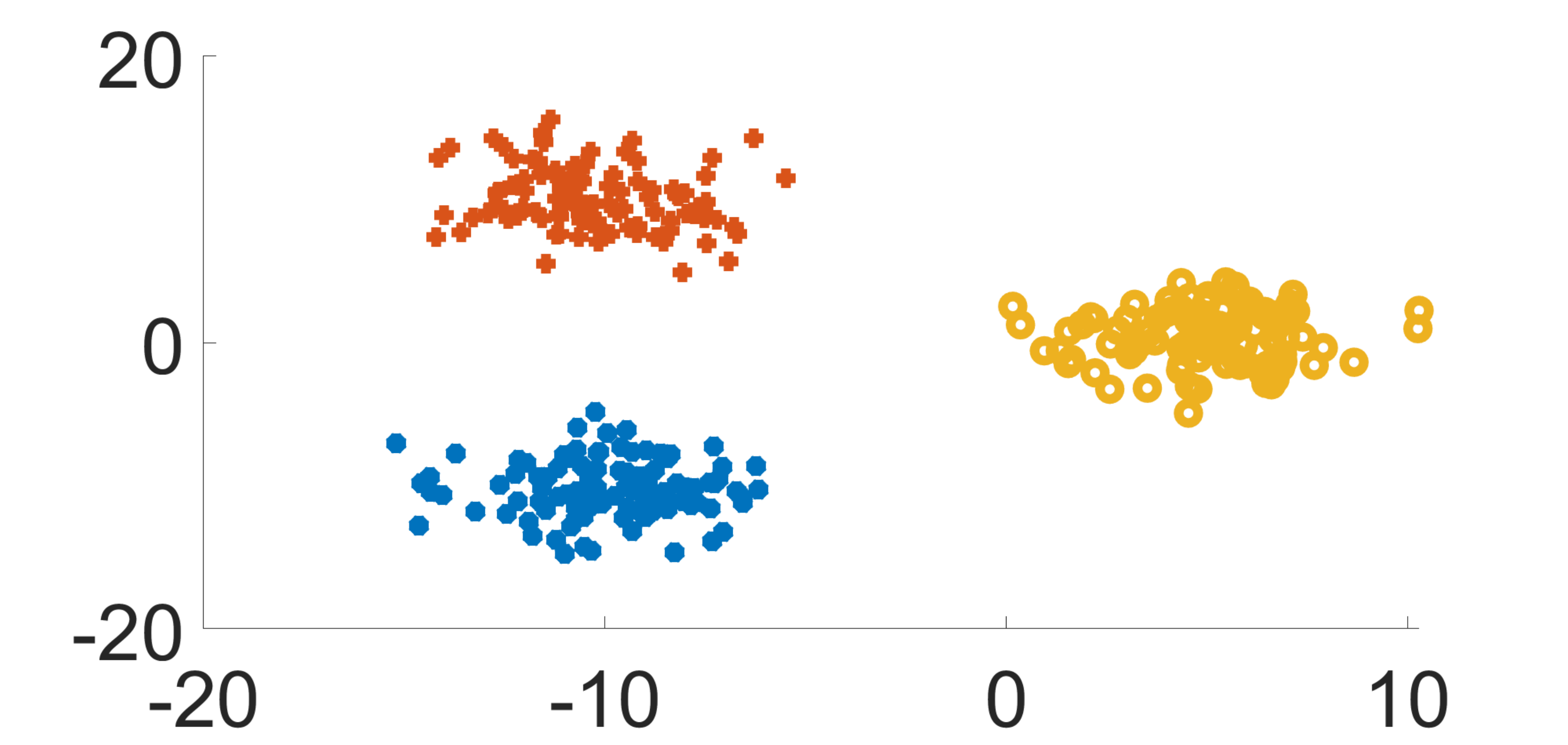}} \hfill
\subfloat[]{\includegraphics[width=0.16\textwidth]{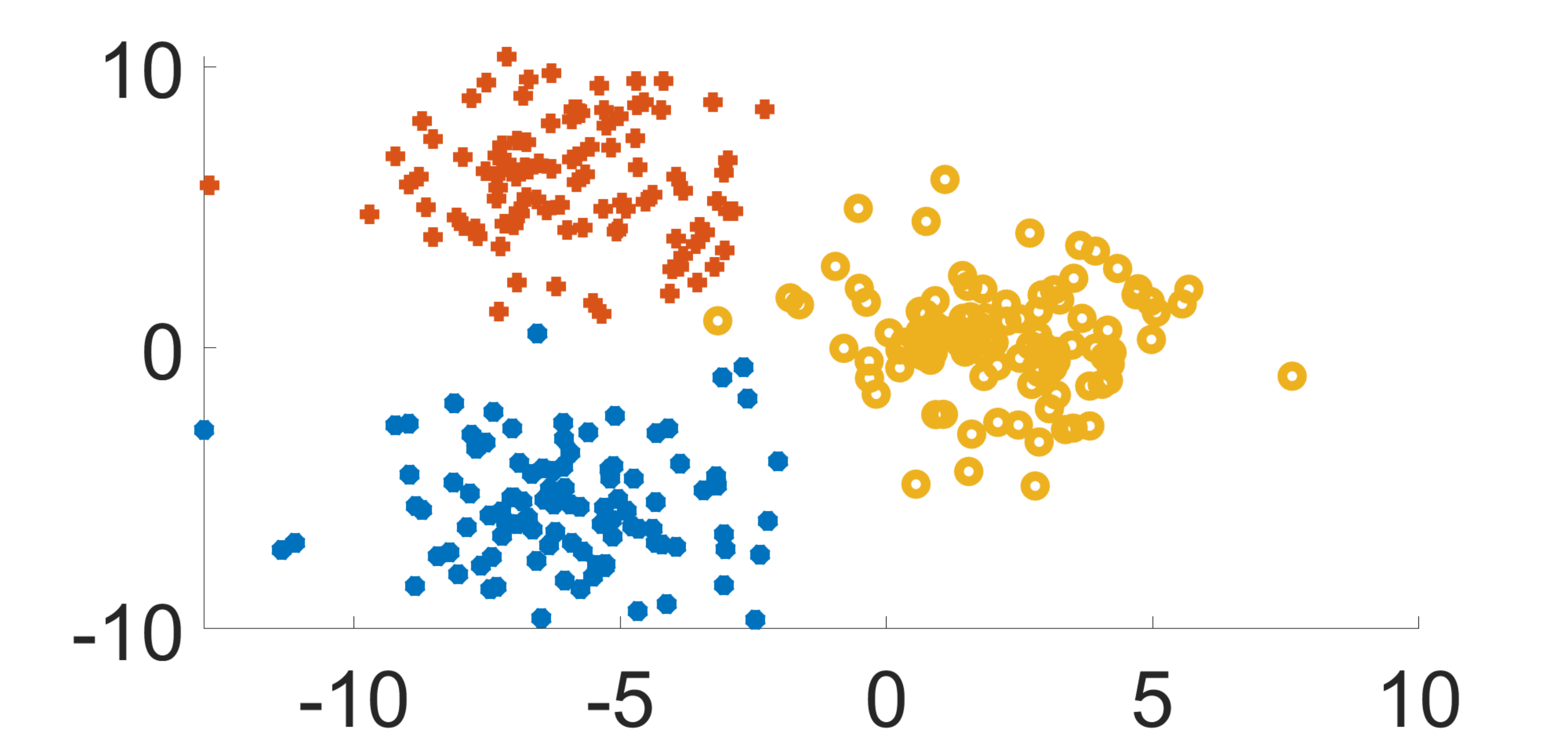}} \hfill
\subfloat[]{\includegraphics[width=0.16\textwidth]{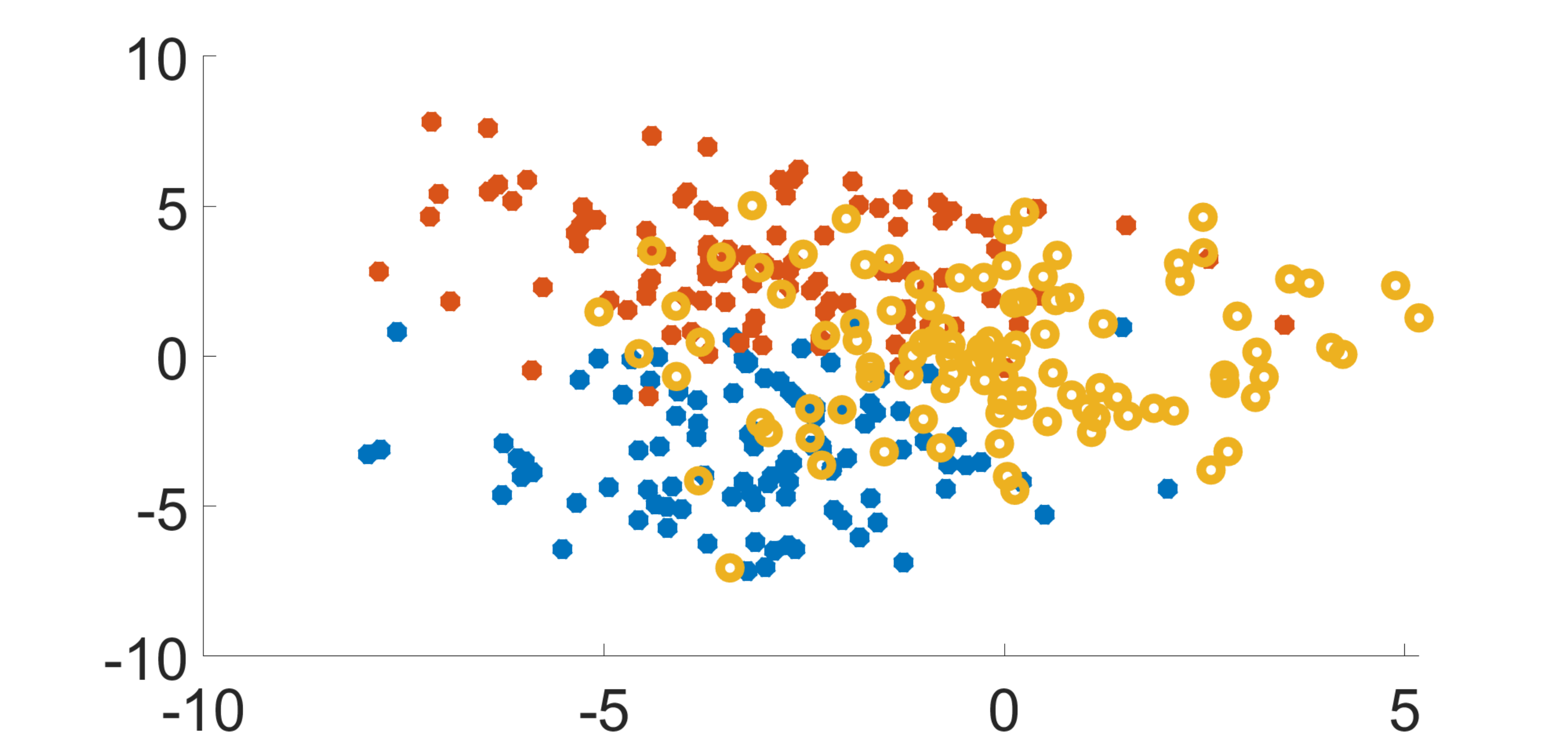}}
\caption{Three different scenarios for clustering.}%
\label{Fig:CS_Non_CS_Dataset}
\end{figure}

\subsection{Constraints Generation and Pre-Processing}
If the partial information available is in the form of labeled data instead of constraints, then the "similar" set, $\mathbb{S}$, is generated by randomly selecting pairs of points from the same class and the "dissimilar" set, $\mathbb{D}$, is generated by randomly selecting pairs of points from different classes. It is important to know that constraints are not very useful in ConiVAT if data has naturally well-separated compact clusters as shown in~\Fig~\ref{Fig:CS_Non_CS_Dataset}(a) because any clustering algorithm can easily detect them without any prior knowledge. Similarly, constraints are of very little use if there is a significantly high overlap between classes, as shown in~\Fig~\ref{Fig:CS_Non_CS_Dataset}(c). 

The instances where constraints will be very beneficial are instances where patterns are at least partially available in the data, but a clustering algorithm would not find them without aid. Such situations arise in many ways in real-data e.g. some points in a cluster may be located far from the dense regions in the same cluster or some points lying near the boundary region of multiple clusters etc.~\Fig~\ref{Fig:CS_Non_CS_Dataset}(b) shows such an example. We have shown in our experiments that ConiVAT works remarkably well for real-data having such cases. 

Constraints provided by domain experts or generated from partially labeled data may be incomplete. However, by using the transitivity property on the available constraints, additional constraints can be generated, as given below
\begin{equation*}
    \mathbf{x}_{1} \xrightarrow{\text{"similar"}} \mathbf{x}_{2} ~ ~ \textrm{and} ~ ~\mathbf{x}_{2} \xrightarrow{\text{"similar"}} \mathbf{x}_{3} ~~ \textrm{then} ~~ \mathbf{x}_{1} \xrightarrow{\text{"similar"}} \mathbf{x}_{3} 
\end{equation*}

\begin{equation*}
       \mathbf{x}_{1} \xrightarrow{\text{"similar"}} \mathbf{x}_{4} ~~  \textrm{and} ~ ~\mathbf{x}_{4} \xrightarrow{\text{"dissimilar"}} \mathbf{x}_{5} ~ \textrm{then} ~ \mathbf{x}_{1} \xrightarrow{\text{"dissimilar"}} \mathbf{x}_{5}
\end{equation*}


The transitive closure expresses the transitive relation between objects. Therefore, we build the transitive closure of the initial constraint sets to expand the constraint sets using Floyd-Warshall algorithm~\cite{cormen2009introduction}. 


Also, in practice, some of the provided constraints can be inconsistent and conflicting. For example,  the "similar" set constraints ($\mathbf{x}_{1}$, $\mathbf{x}_{2}$) and ($\mathbf{x}_{2}$, $\mathbf{x}_{6}$) explicitly conflict with "dissimilar" set ($\mathbf{x}_{1}$, $\mathbf{x}_{6}$). Such inconsistency in constraints can create deadlocks and worsen the performance of our algorithm. Therefore, we remove the inconsistent constraints iteratively to eliminate inconsistency.

\subsection{Integrating Constraints with Metric Learning}
The clusters in the original data space may not be sufficiently separated so distance metric~\cite{xing2003distance}, that satisfies the given constraints, transforms the space to minimize distances between same-cluster datapoints and maximize distances between different-cluster datapoints. Consequently, clusters identified using learned metrics adhere more closely to the concept of similarity embodied in the constraints.

In ConiVAT, the input constraints are used to adapt the underlying distance metric. We parameterize the Euclidean distance using a positive semi-definite matrix $A$ as follows~\cite{xing2003distance}:
\begin{equation}
    d(\mathbf{x}_i, \mathbf{x}_j)= d_A(\mathbf{x}_i, \mathbf{x}_j)= {||\mathbf{x}_i-\mathbf{x}_j||}_{A}= \sqrt{(\mathbf{x}_i-\mathbf{x}_j)^{T} A (\mathbf{x}_i-\mathbf{x}_j)},
\end{equation}
where setting $A=I$ gives Euclidean space and restricting $A$ to be diagonal corresponds to learning a metric in which the different axis are given different "weights"~\cite{xing2003distance}. 

A simple way of specifying a criterion for a desired distance metric that ensures that the points in the "dissimilar" set $\mathbb{D}$ have higher distance between them: ${maximize}_{A}= \sum_{(\mathbf{x}_i, \mathbf{x}_j)\in \mathbb{D}} {||\mathbf{x}_i-\mathbf{x}_j||}_{A}$, and the points in the "similar" set have small squared distance between them . The minimization problem is trivially solved with $A=0$, but adding the constraint $\sum_{(\mathbf{x}_i, \mathbf{x}_j)\in \mathbb{S}} {||\mathbf{x}_i-\mathbf{x}_j||}_{A}^{2} \leq 1$ ensures that $A$ does not collapse the dataset into a single point. The equivalent optimization problem to learn a full matrix $A$  is given as follows~\cite{xing2003distance}:
\begin{eqnarray}
\max_{A} \{ g(A)= \sum_{(\mathbf{x}_i- \mathbf{x}_j)\in \mathbb{D}} {||\mathbf{x}_i,\mathbf{x}_j||}_{A} \\
s.t. \quad \sum_{(\mathbf{x}_i, \mathbf{x}_j)\in \mathbb{S}} {||\mathbf{x}_i-\mathbf{x}_j||}_{A}^{2} \leq 1\\
A\succeq 0\}.
\end{eqnarray}
The choice of constant $1$ on the right hand side in Eq. (3) is arbitrary but not important, and changing it to any other positive constant $c$ would result only in $A$ being replaced by $c^2 A$. The optimization problem in Eqs. (3-5) is convex~\cite{xing2003distance}, so it can be solved  efficiently by gradient ascent and iterative projection~\cite{rockafellar1970convex,dykstra1983algorithm}. The gradient ascent step optimizes $g(A)$ in Eq. (2) and iterative projection ensures that constraints (3) and (4) hold. Precisely, the gradient step $A:= A+\alpha \nabla_{A}g(A)$ and projections of $A$ into sets $C_1 = \{A: \sum_{(\mathbf{x}_i, \mathbf{x}_j)\in \mathbb{S}} {||\mathbf{x}_i-\mathbf{x}_j||}_{A}^{2} \leq 1 \}$ and $C_2 = \{A: A\succeq 0$\} both are repeatedly taken in iterations until termination, as shown in Algorithm~\ref{Algo1}. Termination occurs when
the absolute value of the difference between successive values of the objective function is less than convergence threshold $\epsilon$ or the maximum number of iterations is achieved.

The first projection step in Algorithm~\ref{Algo1} is done by 
solving a sparse set of linear equations, and the second projection step is done by first computing the diagonalization $A= U^T \wedge U $, where $\wedge = diag(\lambda_1, \lambda_2,...,\lambda_N)$ is a diagonal matrix of $A$'s eigenvalues and the columns of $U \in \mathbb{R}^{N \times N}$ contains $A$'s corresponding eigenvectors, and then computing $A^{\prime}= U^T {\wedge}^{\prime} U $ where ${\wedge}^{\prime} =  diag(max(0,\lambda_1), max(0,\lambda_2),...,max(0,\lambda_N))$. Once the full weight matrix $A$ is learned, it is multiplied to the original dataset to obtain the data in the transformed space, and subsequently, to compute the dissimilarity matrix $D$.

\begin{algorithm}
\caption{Metric Learning Algorithm~\cite{xing2003distance}}\label{Algo1}
\begin{algorithmic}
\Statex \textbf{Iterate}
\Statex \quad  \textbf{Iterate}
\Statex \qquad $A:=  \operatorname*{arg\,min}_{A'} \{ {||A'-A ||}_{F}: A' \in C_1 \}$ 
\Statex \qquad $A:=  \operatorname*{arg\,min}_{A'} \{ {||A'-A ||}_{F}: A' \in C_2\}$ \Comment{${||\cdot||}_{F}$ is the Frobenius norm on matrices.}
\Statex \quad  \textbf{A Terminates}
\Statex $A:= A+\alpha{(\nabla_{A}g(A))}$
\Statex \textbf{Termination}

\end{algorithmic}
\end{algorithm}


\subsection{Transitive Dissimilarity Matrix Computation}
Consider dissimilarity matrix $D$ as a transition matrix on a fully connected graph in which each row/column corresponds to a vertex in the graph and each entry in $D$ represents the associated edge weight. Our goal in this step is to alter entries in $D$ so that following intuitions are satisfied even after constraints have been integrated in the metric learning step.

\begin{itemize}
    \item if points $\mathbf{x}_i, \mathbf{x}_j$ are close to each other, then the points that are very near to $\mathbf{x}_i$ are also near to $\mathbf{x}_j$
    
    \item if points $\mathbf{x}_i, \mathbf{x}_j$ are distant from each other, then the points that are very near to $\mathbf{x}_i$ are also distant from $\mathbf{x}_j$
\end{itemize}

In this regard, first we impose the "similar" constraints by reducing the distance between two "similar" points to zero in $D$. By forcing the constraints, we might have violated the triangle inequality. Therefore, we compute a   \textit{minimum transitive dissimilarity} (MTD) matrix $D'$ which respects these new constraints entries in $D$ by running the all-shortest-path algorithm on the modified matrix. Computing the MTD matrix $D'$ from $D$ enhances the cluster structure, while preserving transitivity of the graph. For any path $P_{ij}$ between $\mathbf{x}_i$ and  $\mathbf{x}_j$, the transitive dissimilarity is defined as:
\begin{eqnarray}
T(P_{ij})=   \displaystyle \operatorname*{max}_{P_{ij}} (d_{i,k_1}, d_{k_1,k_2},...,d_{k_{n-1},k_n}, d_{k_n,k_j})
\end{eqnarray}
Then, among all existing paths between $\mathbf{x}_i$ and  $\mathbf{x}_j$, the MTD is defined as:
\begin{eqnarray}
d_{ij}^{\prime}=  \displaystyle \operatorname*{min}_{p \in P_{ij}} T(P_{ij}),
\end{eqnarray}
which implies that dissimilarity between two objects is defined by the largest edge weight (lowest density) on the minimal connecting path between two objects. Computing the MTD matrix $D'$ directly from $D$ using Eqs. (5) and (6) 
is equivalent to the path based distance transformation~\cite{fischer2001path} mentioned in the non-recursive version of iVAT~\cite{wang2010ivat}. The path-based minimax distance transform significantly improves the assessment of cluster structure by reducing the distances among the same clusters (diagonal dark blocks in the RDI), while keeping the distance between two clusters fixed. 

In the final step, VAT is applied on the MTD matrix $D'$ which returns a reordered matrix ${D'}^{*}$ and the cut magnitudes of MST links. The visualization of ${D'}^{*}$ suggests the number of clusters $k$ present the dataset. Having the estimate of the number of clusters, $k$, from visual observation of the ${D'}^{*}$, we cut the $k-1$ longest edges in the MST, resulting in $k$ single-linkage clusters. 

\section{Experiments and Results}\label{sec:experiments}
In this section, we perform several experiments on nine datasets to evaluate our proposed ConiVAT algorithm. In the first experiment, we compare images from iVAT and its semi-supervised version, ConiVAT, to gauge the quality of reordered dissimilarity images (RDI) for cluster tendency assessment and clustering performance on all nine datasets. In the second experiment, we compare ConiVAT with two unsupervised and three constraint-based hierarchical clustering approaches, based on the clustering performance. In the third experiment, we evaluate the improvements from each component in ConiVAT.  In the last experiment, we examine the effects of constraints in the ConiVAT framework.

\subsection{Datasets}
Table~\ref{table:Datasets} lists the nine datasets used in our experiments. SYNTH1 and SYNTH2 are 2D synthetic datasets, and the remaining seven are real-datasets that are publicly available at the UCI Machine Learning database repository~\cite{blake1998uci}.  All the datasets are labeled. We point out that the labeled subsets in these data sets may or may not correspond to computationally identifiable sets of clusters. MUSH and SHUTTLE are subsets of corresponding full datasets, having balanced class distributions. Note that the labels were used only for evaluation and constraints generation in our experiments. All real datasets were pre-processed by removing the missing data and normalizing the features between $0$ and $1$\footnote{The features are normalized to the interval [0,1] by subtracting the minimum and then dividing by the subsequent maximum so that they all had the same scale.}.

\begin{table}[]
\caption{Datasets Descriptions}
\label{table:Datasets}
\resizebox{0.5\textwidth}{!}{%
\begin{tabular}{cccc}
\hline
\multicolumn{1}{|c|}{\textbf{Dataset}} & \multicolumn{1}{c|}{\textbf{\# Samples}} & \multicolumn{1}{c|}{\textbf{\# Dimensions}} & \multicolumn{1}{c|}{\textbf{\#Classes}} \\ \hline
\multicolumn{1}{|c|}{SYNTH1} & \multicolumn{1}{c|}{400}  & \multicolumn{1}{c|}{2}  & \multicolumn{1}{c|}{4} \\ \hline
\multicolumn{1}{|c|}{SYNTH2} & \multicolumn{1}{c|}{750}  & \multicolumn{1}{c|}{2}  & \multicolumn{1}{c|}{3} \\ \hline
\multicolumn{1}{|c|}{IRIS}      & \multicolumn{1}{c|}{150}  & \multicolumn{1}{c|}{4}  & \multicolumn{1}{c|}{3} \\ \hline
\multicolumn{1}{|c|}{WINE}      & \multicolumn{1}{c|}{178}  & \multicolumn{1}{c|}{13} & \multicolumn{1}{c|}{3} \\ \hline
\multicolumn{1}{|c|}{VOTING}    & \multicolumn{1}{c|}{435}  & \multicolumn{1}{c|}{16} & \multicolumn{1}{c|}{2} \\ \hline
\multicolumn{1}{|c|}{MUSH}  & \multicolumn{1}{c|}{1000} & \multicolumn{1}{c|}{22} & \multicolumn{1}{c|}{2} \\ \hline
\multicolumn{1}{|c|}{PIMA}      & \multicolumn{1}{c|}{768}  & \multicolumn{1}{c|}{8}  & \multicolumn{1}{c|}{2} \\ \hline
\multicolumn{1}{|c|}{SHUTTLE}      & \multicolumn{1}{c|}{400}  & \multicolumn{1}{c|}{10}  & \multicolumn{1}{c|}{4} \\ \hline
\multicolumn{1}{|c|}{SOYBEAN}      & \multicolumn{1}{c|}{562}  & \multicolumn{1}{c|}{35}  & \multicolumn{1}{c|}{19} \\ \hline
\multicolumn{1}{l}{}            & \multicolumn{1}{l}{}      & \multicolumn{1}{l}{}    & \multicolumn{1}{l}{}  
\end{tabular}%
}
\end{table}

\subsection{Evaluation Criteria}
In the comparative study between ConiVAT and iVAT, both were evaluated based on the quality of the output RDI in estimating the number of clusters, including cluster structure and clustering performance. Similar to the VAT/iVAT algorithms, SL-partitions are aligned partitions corresponding to ConiVAT reordered dissimilarity matrices. 

For all datasets, the quality of the output crisp partition is assessed using ground truth information.  The similarity of computed partitions to ground-truth labels is measured using the \textit{partition accuracy} (PA). The PA of a clustering algorithm is the ratio of the number of objects with matching ground truth labels and algorithmic labels to the total number of objects in the data. The value of (\%)PA ranges from $0$ to $100$, and a higher value implies a better match to the ground truth partition. Before the PA can be calculated, we ensure that the algorithmic labels obtained from the clustering algorithms correspond to the same subsets in the ground truth. We also report the run-time (in seconds) for ConiVAT and baseline algorithms compared in this work.

\subsection{Computation Protocols and Parameter Settings}
All the experiments were performed using MATLAB on a Windows 10 Enterprise (64bit) with 16GB RAM and Intel i7@1.90GHz Processor. To generate constraints for each data, we randomly select two instances from the data and check their labels (which are used only for evaluation  and not for clustering). If they are from the same class, we put the pair in the "similar" constraint set, else in the "dissimilar" constraint set. For all the datasets, we generate $30$ constraints, unless stated otherwise. Since constraints are generated randomly, the reported results in our experiments are computed by averaging $10$ runs for each semi-supervised algorithm with $2$-fold cross-validations\footnote{The ConiVAT RDI presented in our first experiment is selected randomly from  $10$ RDIs (obtained from $10$ runs) for each dataset.}. The ConiVAT, SSL, and CCL algorithm consider constraints in the same form i.e., "similar" and "dissimilar" constraints; however, UltraTran considers "triplet" constraints in its framework. Therefore,  to have a fair comparison, we generate triplet constraints $\{\mathbf{x}_{a},\mathbf{x}_{b},\mathbf{x}_{c} \}$ where $\mathbf{x}_{a},\mathbf{x}_{b}  \in \textrm{Class~ i} $  (taken from "similar" set constraints) and $\mathbf{x}_{c} \in \textrm{Class~j} (\neq i)$ for the UltraTran algorithm.  Note that the input constraints for semi-supervised algorithms may not have the representations from all known classes of a dataset.

In ConiVAT, we learned a full weight matrix,$A$, using metric-learning. The convergence threshold = $0.001$, the number of maximum iterations = $100$, gradient rate $\alpha= 0.1$, and the number of maximum iterative projections is chosen as $10000$ for coniVAT. All the algorithms in this work except iVAT and ConiVAT require the number of clusters as an input to obtain output clustering partition for each dataset. However,  since iVAT and ConiVAT provide cluster tendency assessment, they do not require the target number of clusters to be provided explicitly as an input. The Euclidean distance was considered as the default metric in all models. 

\subsection{Comparison of iVAT and ConiVAT}
In this experiment, we compare the iVAT and ConiVAT algorithm based on the quality of RDI to estimate the number of clusters in the data and (\%)PA for clustering performance. 

\begin{figure*}
\captionsetup[subfigure]{justification=centering}
\centering
\subfloat[$k=4$]{\includegraphics[width=0.3\textwidth, height= 0.16\textheight]{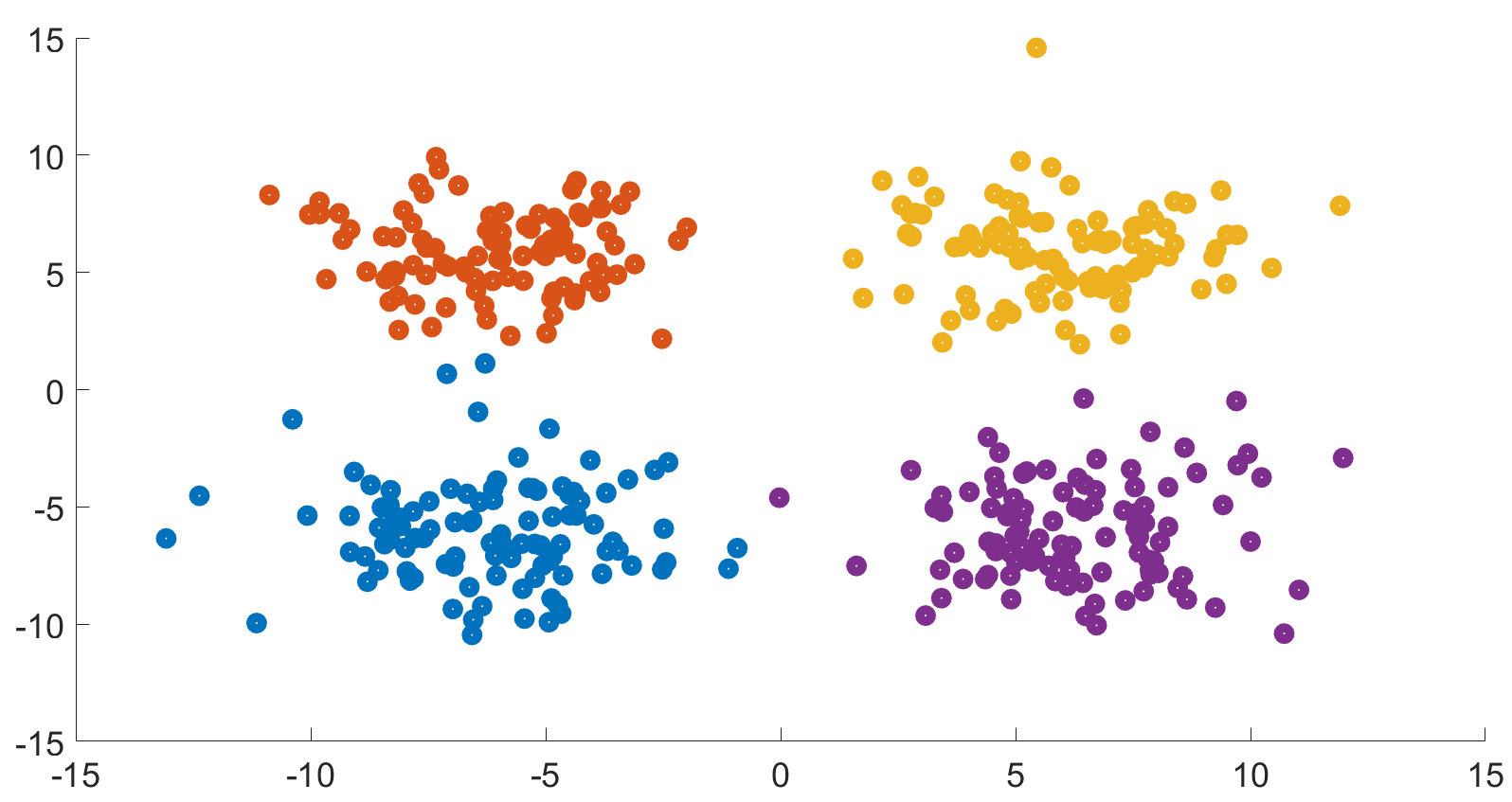}} \hfill
\subfloat[PA: 55.2\%]{\includegraphics[width=0.2\textwidth]{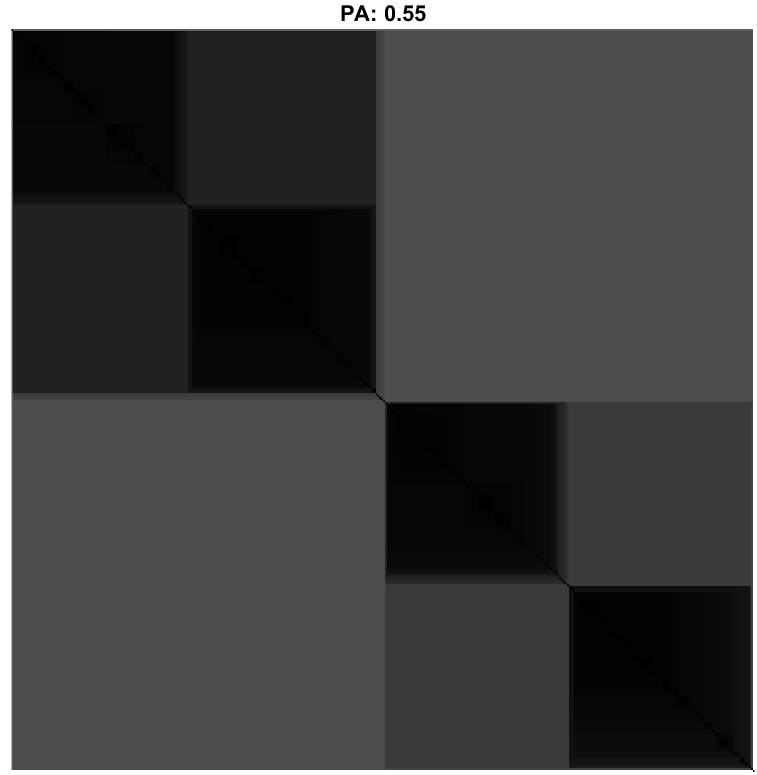}} \hfill
\subfloat[PA: 92.3\%]{\includegraphics[width=0.2\textwidth]{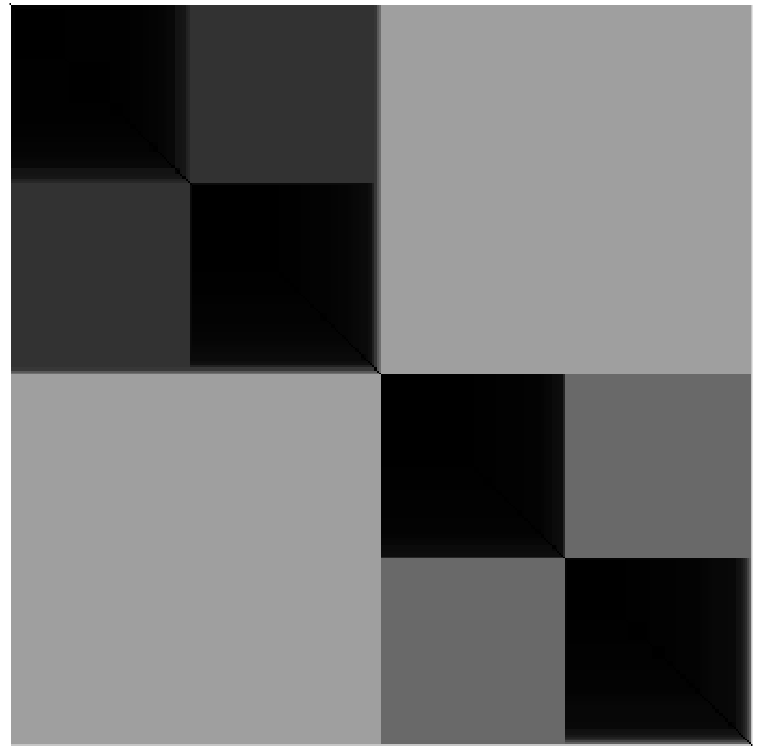}}
\caption{(a) SYNTH1: 2D Synthetic dataset (b) iVAT image (PA: 55.2\%); (c) ConiVAT image (PA: 92.3\%).}%
\label{Fig:2D_Dataset}
\end{figure*}

Our first example is a 2D synthetic dataset, SYNTH1, constructed by drawing labeled samples from a mixture of $k=4$ Gaussian distributions, as shown in \Fig~\ref{Fig:2D_Dataset} (a). It can be seen that SYNTH1 dataset has some inliers between clusters. Also, there are some points in a few clusters (e.g., blue and yellow datapoints), far away from their dense regions, that may ruin the iVAT image and deteriorate clustering performance. The corresponding iVAT image in \Fig~\ref{Fig:2D_Dataset} (b) shows three (hard to see) dominant dark blocks along its diagonal, suggesting that there are $3$ clusters in the dataset, which is an incorrect estimate of $k$. The iVAT image also reflects the far away points and inliers of SYNTH1 dataset in its centre as a few singleton dark blocks that could have ruined the iVAT image, thus presenting an incorrect number of clusters and $55.2\%$ PA for clustering. The ConiVAT image for this dataset always (in all 10 runs) displays four dark blocks (in View (c), the two at the upper left are really there) along its diagonal corresponding to the four clusters present in the SYNTH1 dataset, and achieves $92.3\%$ clustering accuracy. This indicates that by integrating some constraints in iVAT using the ConiVAT framework significantly improves the quality of the RDI and SL clustering performance.

\begin{figure*}
\captionsetup[subfigure]{justification=centering}
\centering
\subfloat[$k=3$]{\includegraphics[width=0.3\textwidth, height= 0.16\textheight]{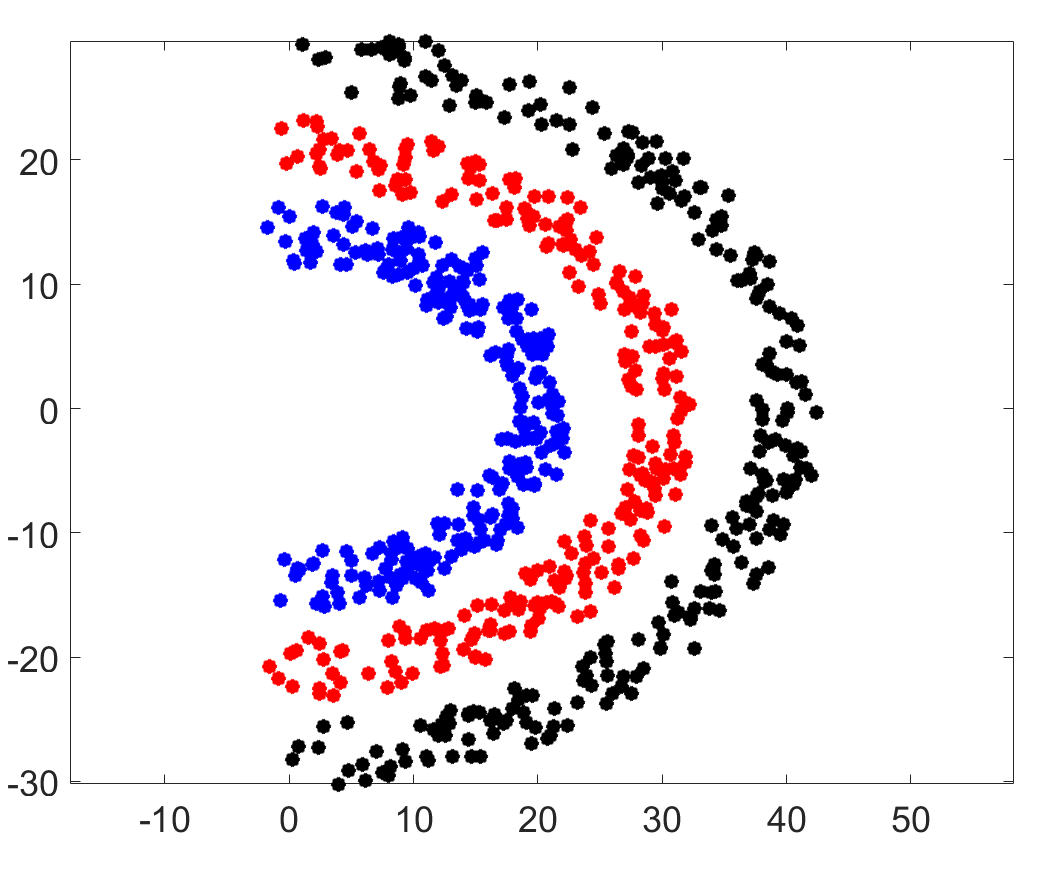}} \hfill
\subfloat[PA: 34.9\%]{\includegraphics[width=0.2\textwidth]{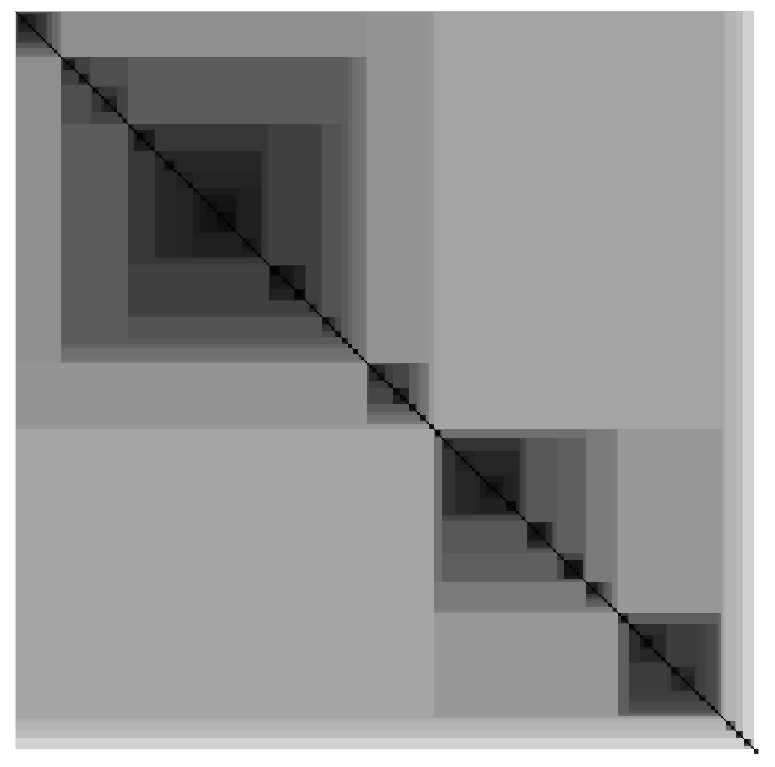}} \hfill
\subfloat[PA: 78.3\%]{\includegraphics[width=0.2\textwidth]{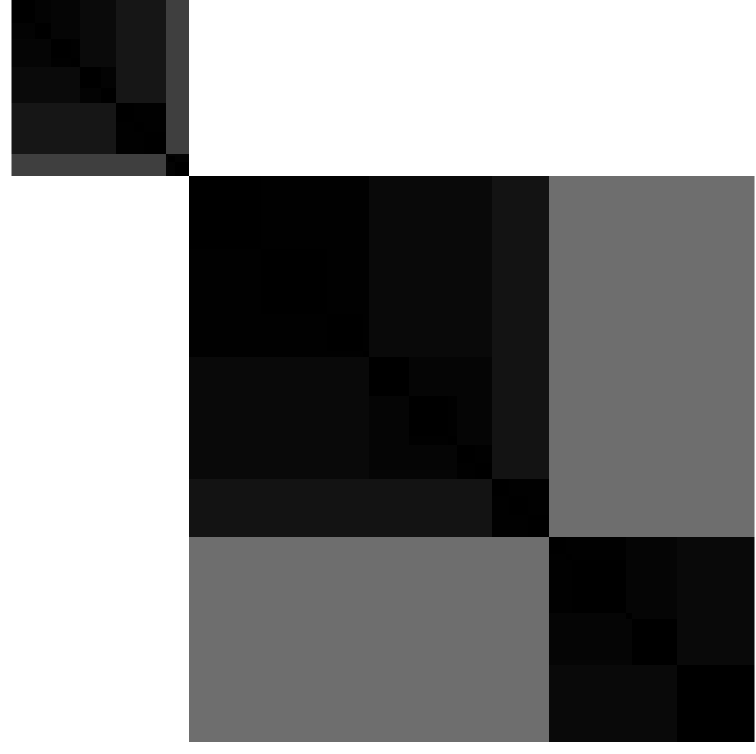}}
\caption{(a) SYNTH2: Banana dataset (b) iVAT image (PA: 34.9\%); (c) ConiVAT image  (PA: 78.3\%).}%
\label{Fig:Half_Kernel_Dataset}
\end{figure*}

SYNTH2 is a 2D synthetic dataset which has three different semi-ellipsoidal clusters of equal sizes but they are not well-separated. The corresponding iVAT image in \Fig~\ref{Fig:Half_Kernel_Dataset} (a) shows 1 big,  2 middle-size, and 5-6 tiny dark blocks suggesting that there may be $3$ dominant clusters but they are connected by some noisy points or inliers. Consequently, SL clustering using iVAT achieves only $34.9\%$ PA. On the other hand, the ConiVAT image in \Fig~\ref{Fig:Half_Kernel_Dataset} (b) shows three dominant dark blocks confirming three clusters in the SYNTH2 dataset. However, the size of the dark blocks are not proportional (2 equal-sized small dark blocks and 1 big dark block) to the actual number of datapoints in each cluster, therefore, it achieves $78.3\%$ average clustering accuracy.

\begin{figure}
\captionsetup[subfigure]{justification=centering}
\centering
\subfloat[]{\includegraphics[width=0.23\textwidth]{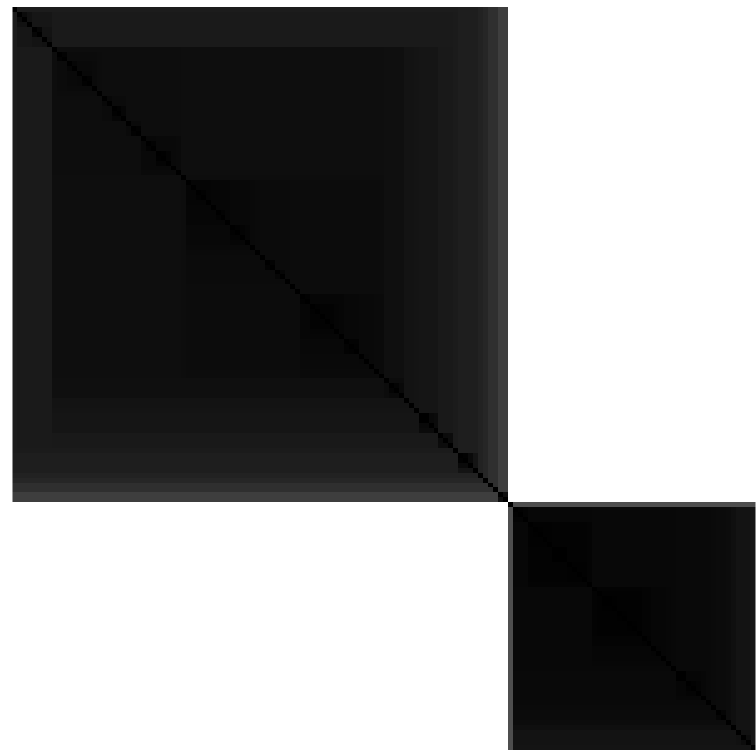}}
\subfloat[]{\includegraphics[width=0.23\textwidth]{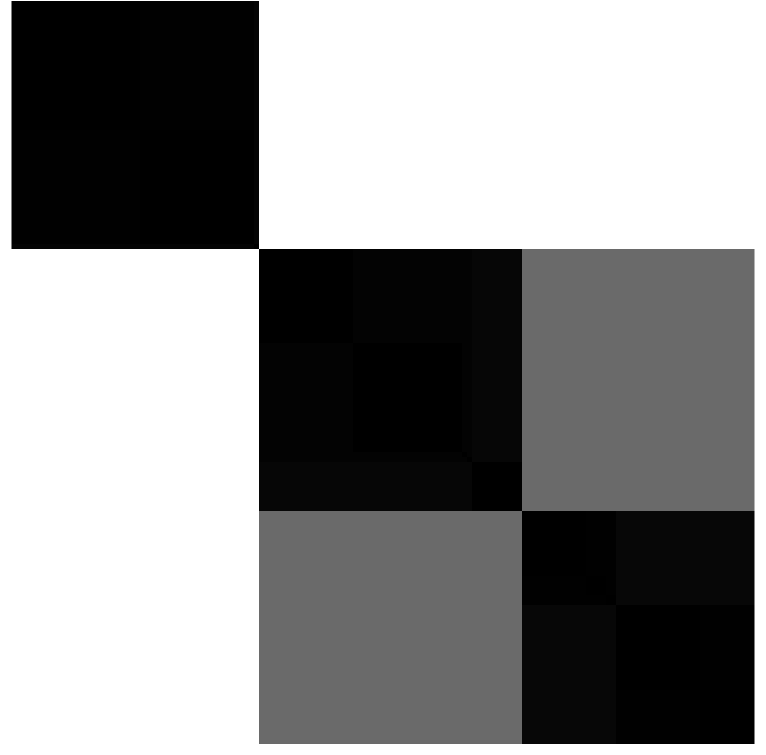}}
\caption{IRIS dataset (a) iVAT image (PA: 66\%); (b) ConiVAT image (PA: 98\%).}%
\label{Fig:IRIS_Dataset}
\end{figure}

Our third example is on the IRIS dataset which has three classes (flower types) that are not well-separated and are probably not hyper-spherical in their 4D input space.~\Fig~\ref{Fig:IRIS_Dataset} (a) shows the corresponding iVAT image displaying one big dark block and one small block that suggests there are two clusters in the IRIS dataset. Consequently, it achieves only $66\%$ clustering accuracy. This is probably because two of three flower types in IRIS data greatly overlap in $\mathbb{R}^{4}$ so it is often argued that naturally, it has only $2$ well-separated clusters. In contrast, ConiVAT clearly displays three equally-sized dark blocks along its diagonal (view (b)) suggesting three equal-sized SL clusters in IRIS dataset, thus achieving significant improvement in clustering accuracy (98\%) over iVAT. 

\begin{figure}
\captionsetup[subfigure]{justification=centering}
\centering
\subfloat[]{\includegraphics[width=0.23\textwidth]{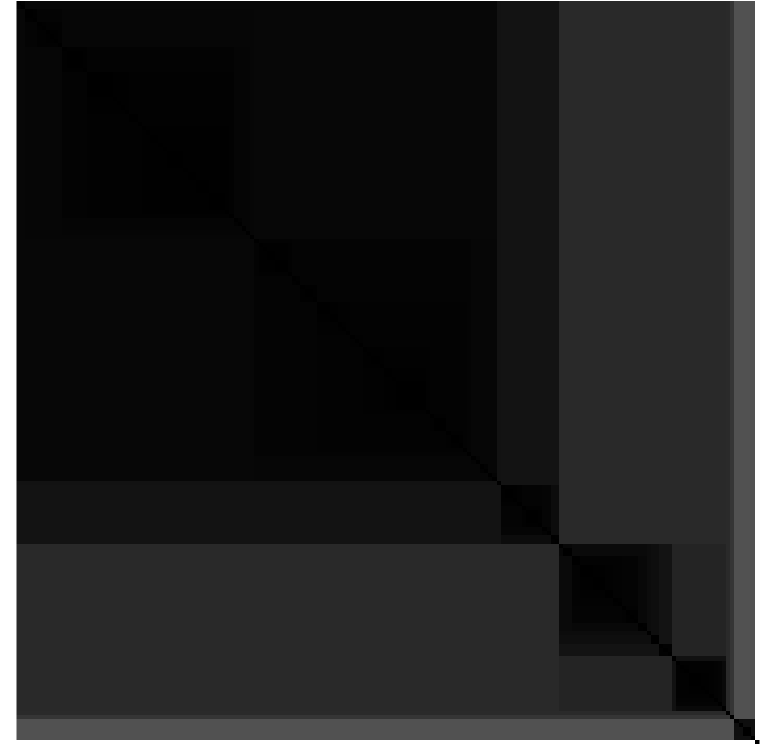}}
\subfloat[]{\includegraphics[width=0.23\textwidth]{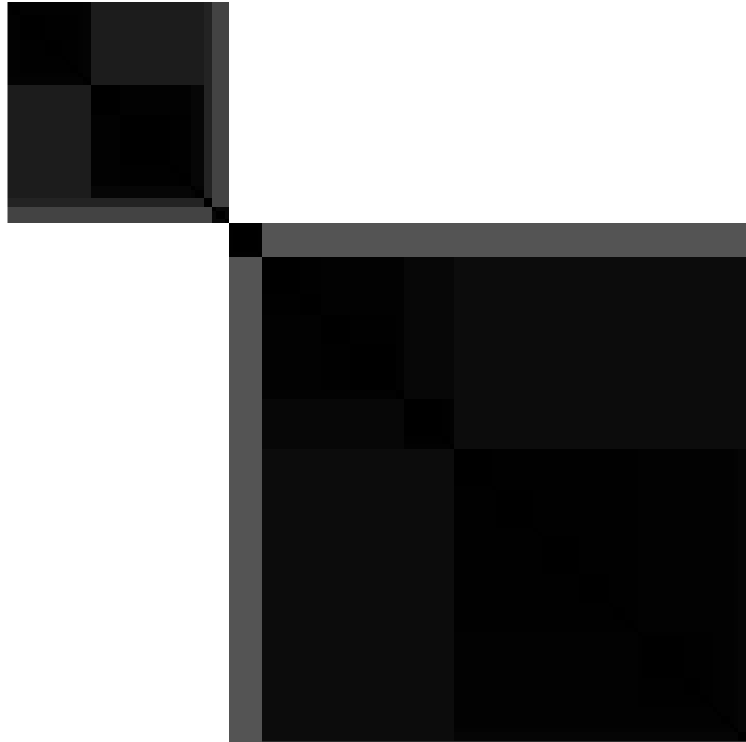}}
\caption{WINE dataset (b) iVAT image (PA: 42\%); (b) ConiVAT image (PA: 66.4\%).}%
\label{Fig:Wine_Dataset}
\end{figure}

The fourth example is on the $13$D WINE dataset, which also apparently has $3$ overlapping classes (types of wines). The corresponding iVAT image in~\Fig~\ref{Fig:Wine_Dataset} (a) shows one big dark block, comprising three dominant sub-blocks within it, and one tiny dark block at the right bottom. This suggests that there may be an outlier in the WINE dataset, corresponding to the tiny dark block, which is far away from the three normal clusters. Consequently, the three normal clusters appear as one big cluster and an outlier appears as a tiny cluster in the iVAT image, thus SL clustering using iVAT achieves $42\%$ accuracy. On the other hand, ConiVAT clearly shows three main dark blocks in~\Fig~\ref{Fig:Wine_Dataset} (b) suggesting three clusters in the WINE dataset, and achieves $66.4\%$ clustering accuracy.

\begin{figure}
\captionsetup[subfigure]{justification=centering}
\centering
\subfloat[]{\includegraphics[width=0.23\textwidth]{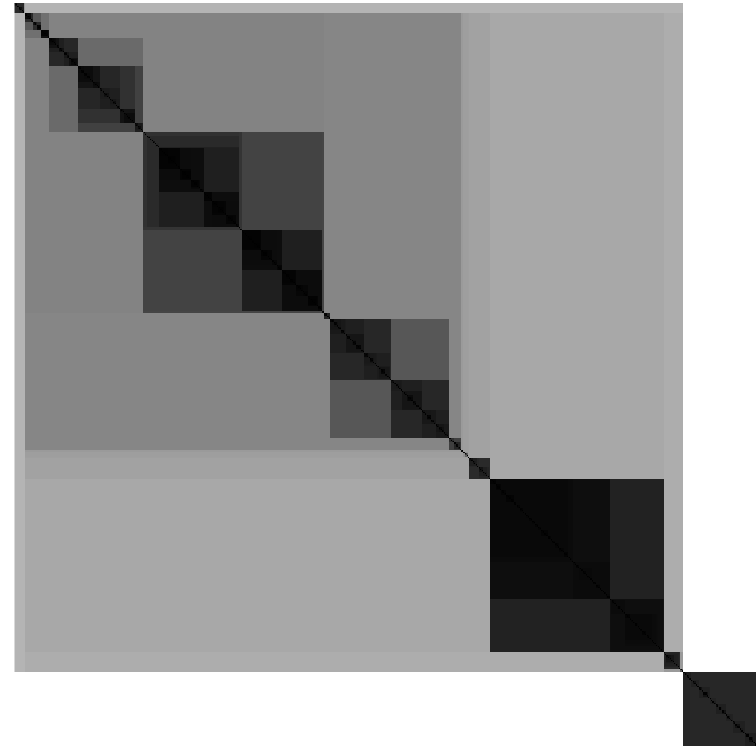}}
\subfloat[]{\includegraphics[width=0.23\textwidth]{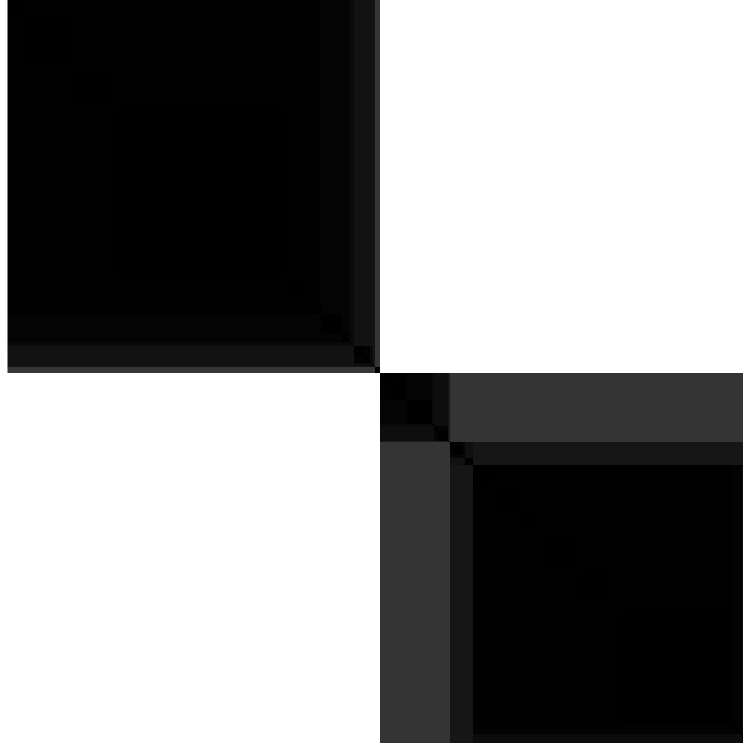}}
\caption{MUSH dataset; (a) iVAT image (PA: 58.1\%); (b) ConiVAT image (PA: 97.3\%).}%
\label{Fig:Mushroom_Dataset}
\end{figure}

The MUSH dataset contains $1000$ randomly selected points from the full mushroom data, and it has two classes: poisonous or edible. The corresponding iVAT image in ~\Fig~\ref{Fig:Mushroom_Dataset} (a) shows more than $10$ dark blocks of varying size, suggesting more than $10$ clusters in the data. On the other hand, the corresponding ConiVAT image in~\Fig~\ref{Fig:Mushroom_Dataset} (b) shows two dominant dark blocks suggesting the actual number of clusters ($2$) present in the data. As a result, ConiVAT achieves $97.3\%$ clustering accuracy, compared to $58.1\%$ accuracy from iVAT.

\begin{figure}
\captionsetup[subfigure]{justification=centering}
\centering
\subfloat[]{\includegraphics[width=0.23\textwidth]{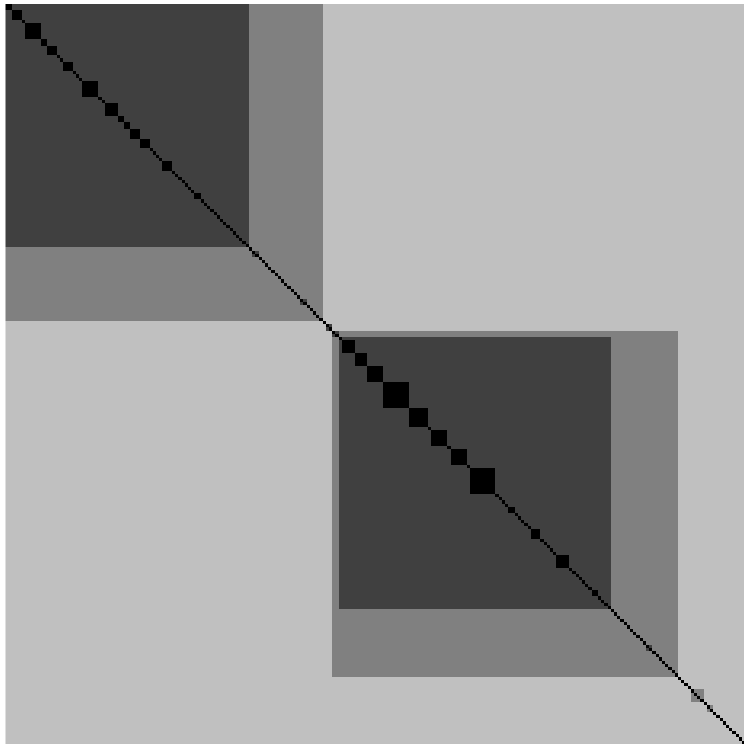}}
\subfloat[]{\includegraphics[width=0.23\textwidth]{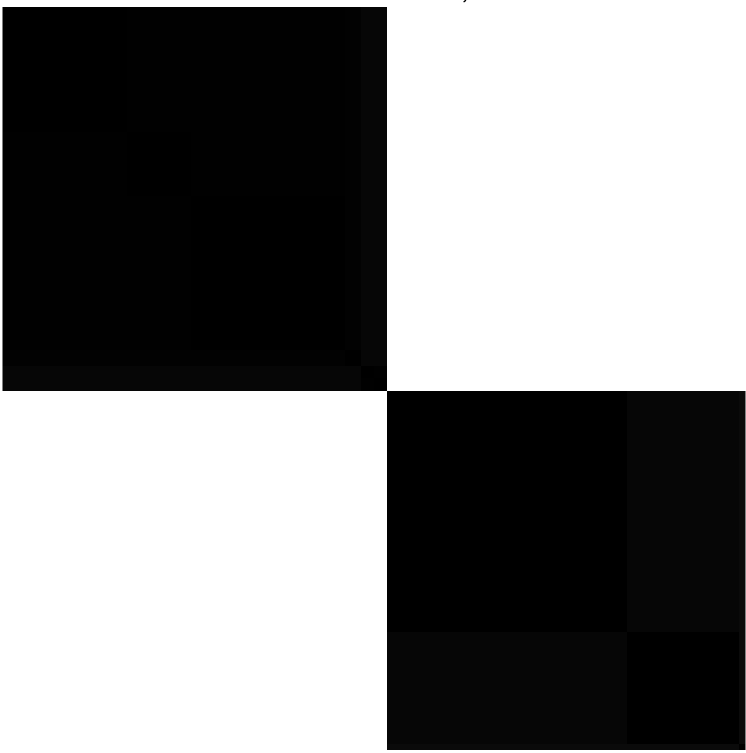}}
\caption{VOTING dataset; (a) iVAT image (PA: 60.2\%); (b) ConiVAT image (PA: 95.2\%).}%
\label{Fig:Voting_Dataset}
\end{figure}

The next example is on the VOTING dataset which includes votes of $435$ members of the U.S. House of Representative on $16$ key issues in $1994$ with three different types of votes: yes, no, or undecided. This data (Representative's Party) is labeled as a Democrat or Republican. The presence of a two dominant dark block in the corresponding iVAT image in~\Fig~\ref{Fig:Voting_Dataset} (a) indicates two clusters in the data. However, there are many singleton clusters (tiny dark blocks) as inliers and outliers that restrict the clustering accuracy of iVAT to $60.2\%$. The ConiVAT image in~\Fig~\ref{Fig:Voting_Dataset} (b) clearly displays two dark blocks indicating two clusters in the VOTING dataset, and achieves significant improvement in clustering with $95.2\%$ accuracy.

\begin{figure}
\captionsetup[subfigure]{justification=centering}
\centering
\subfloat[]{\includegraphics[width=0.23\textwidth]{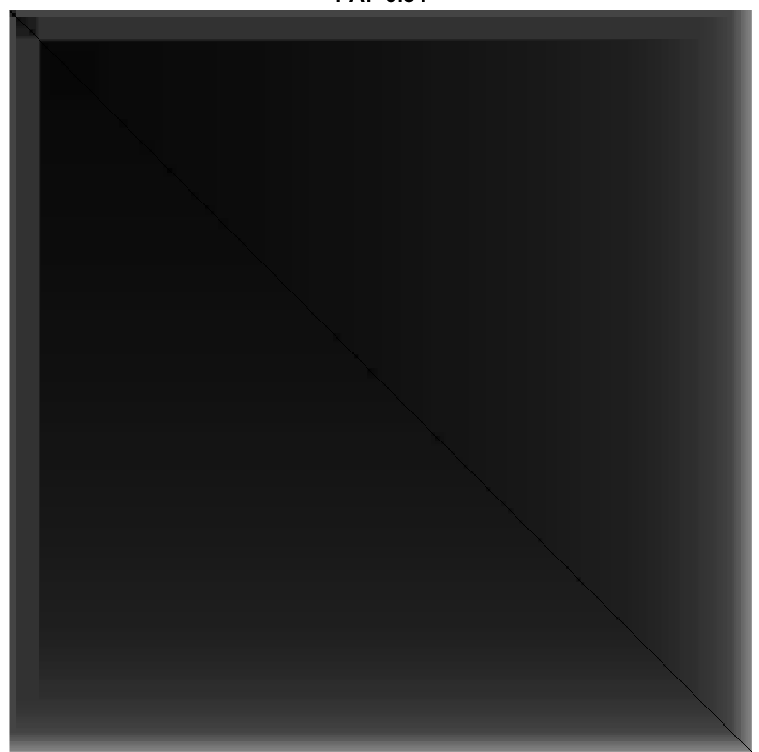}}
\subfloat[]{\includegraphics[width=0.23\textwidth]{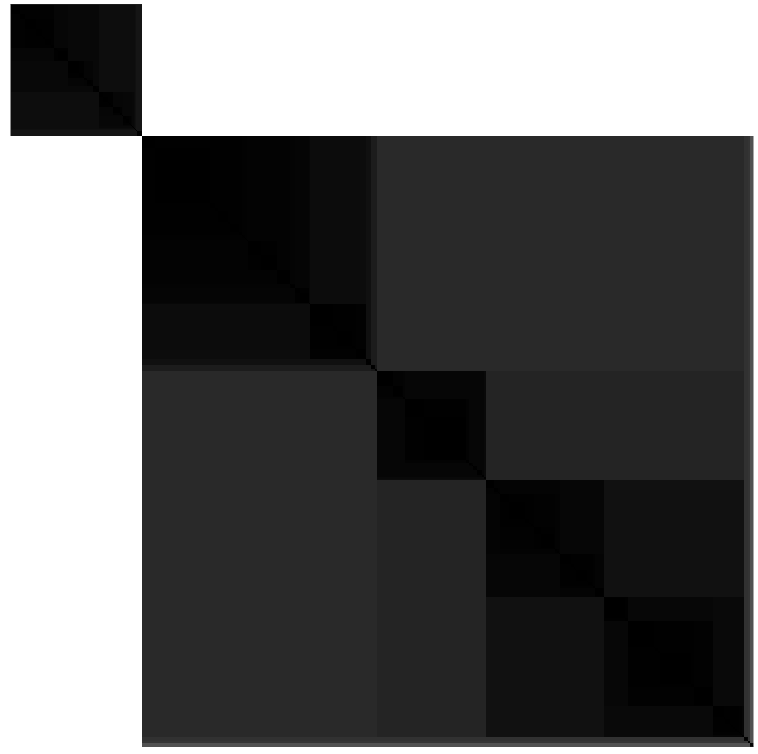}}
\caption{PIMA dataset; (a) iVAT image (PA: 34.8\%); (b) ConiVAT image (PA: 37.3\% for k=3, 66.7\% for k=4).}%
\label{Fig:Diabetes_Dataset}
\end{figure}

PIMA is a Pima Indian Diabetes dataset which contains $786$ samples and $8$ attributes, labeled in two classes: diabetic and healthy individuals. The corresponding iVAT image in~\Fig~\ref{Fig:Diabetes_Dataset}(a) shows a single dark block indicating a single cluster in the dataset, and it achieves $34.8\%$ clustering accuracy. The ConiVAT image in~\Fig~\ref{Fig:Diabetes_Dataset}(b) indicates on one small dark block on the top left and one big dark block on the right bottom of the image. Upon a much closer look at the big dark block, one can see three sub-blocks within it suggesting that on a coarse level there are two clusters; however, on a finer level, they may be four clusters in the data. Consequently, the ConiVAT achieves $37.9\%$ clustering accuracy for $k=2$ and $66.7\%$ for $k=4$ for PIMA dataset. This example shows that even though constraints are generated from all the available classes in the data ($2$ in this data), ConiVAT can reflect other patterns in the data.

\begin{figure}
\captionsetup[subfigure]{justification=centering}
\centering
\subfloat[]{\includegraphics[width=0.23\textwidth]{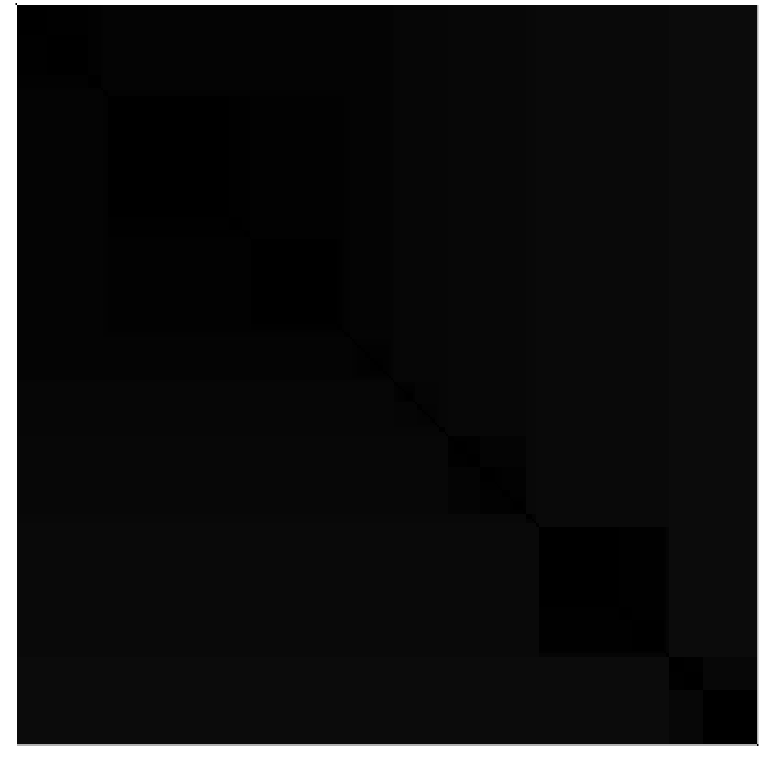}}
\subfloat[]{\includegraphics[width=0.23\textwidth]{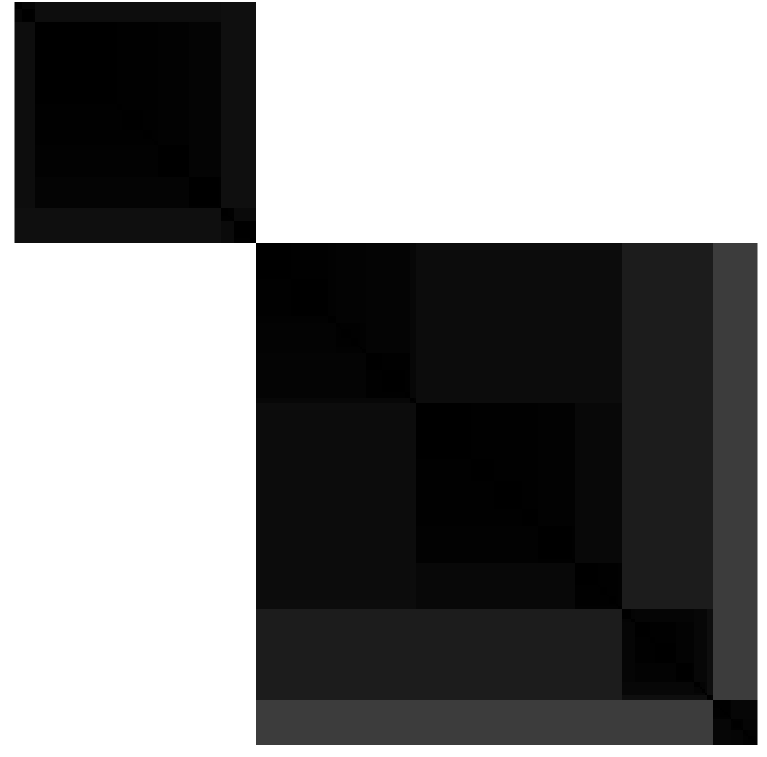}}
\caption{Shuttle dataset; (a) iVAT image (PA: 25.3\%); (b) ConiVAT image  (PA: 39.1\%).}%
\label{Fig:Shuttle_Dataset}
\end{figure}

The SHUTTLE dataset is a subset of the full Statlog (Shuttle) dataset, which has $14,500$ records with $9$ attributes, labeled in the $9$ classes. The classes 1, 3, 4 and 5 are the most dominant classes that account for 99.6\% of all samples available; therefore, we randomly select an equal number of samples from the full dataset to construct a balanced dataset, SHUTTLE. The corresponding iVAT image in~\Fig~\ref{Fig:Shuttle_Dataset} (a) shows a single dark block indicating only a single cluster in the SHUTTLE dataset. On the other hand, the ConiVAT image shows $2$ dark blocks on low-resolution, however, four dark blocks on high-resolution confirms $4$ clusters in the dataset. The ConiVAT achieves $39.1\%$ clustering accuracy compared to $25.3\%$ accuracy of the iVAT.

\begin{table*}[]
\caption{Comparison of two unsupervised and four semi-supervised clustering algorithms based on the PA (\%). }
\label{table:Comparison}
\resizebox{\textwidth}{!}{%
\begin{tabular}{|c|c|c|c|c|c|c|c|c|c|}
\hline
\textbf{Dataset} & \textbf{SYNTH1} & \textbf{SYNTH2} & \textbf{IRIS} & \textbf{WINE} & \textbf{VOTING} & \textbf{MUSH} & \textbf{PIMA} & \textbf{SHUTTLE} & \textbf{SOYBEAN} \\ \hline
ConiVAT & 92.3 & \textbf{78.3} & \textbf{98.0} & \textbf{66.4} & \textbf{95.2} & \textbf{97.3} & \textbf{66.7} &  \textbf{39.1}& \textbf{46.3} \\ \hline
UltraTran      & 75.3 &65.3 & 66.7 & 65.3 & 61.6 &   57.5&  34.9 & 25.8 &   40.9\\ \hline
SSL            &   74.9   & 48.9      &    67.8       &   45.8        &      61.2     &     67.8      &   34.9        & 25.5  & 41.2 \\ \hline
CCL            &    \textbf{93.9} & 36.9    &      86.7       &   60.8        &       61.2   &     65.7      &   35.3        & 33.5  &  43.2\\ \hline
HAC-SL         &   74.5   & 45.9     &  66.0         &   42.4       &     61.6      &     73.3      &      34.9     & 25.3 &  40.3 \\ \hline
HAC-CL         &    92.8   & 36.5   &   90.7        &   60.6        &   61.7        &       69.8    &     34.9      &   26.5  & 41.4  \\ \hline
\end{tabular}%
}
\end{table*}

\begin{figure}
\captionsetup[subfigure]{justification=centering}
\centering
\subfloat[]{\includegraphics[width=0.23\textwidth]{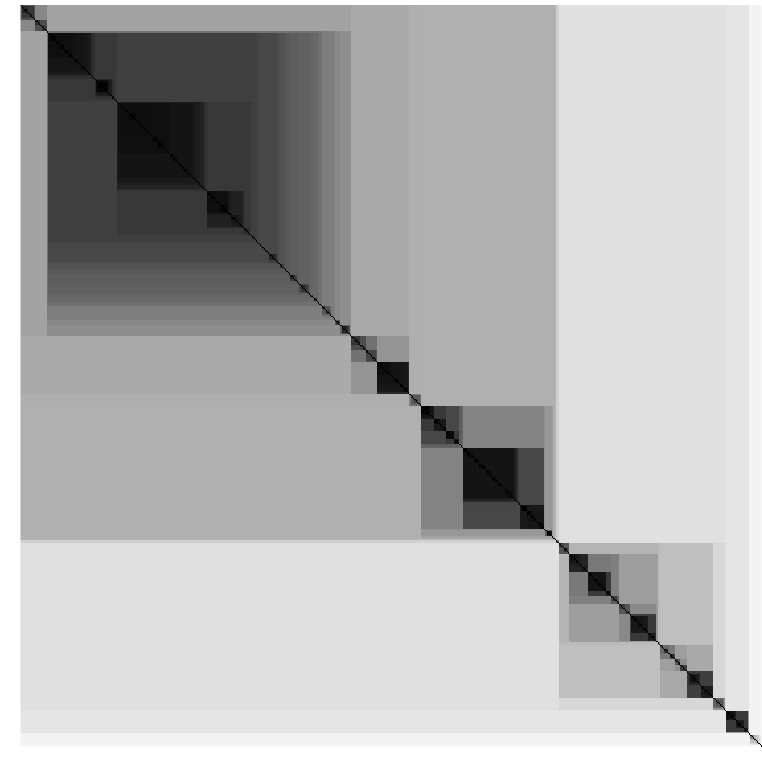}}
\subfloat[]{\includegraphics[width=0.23\textwidth]{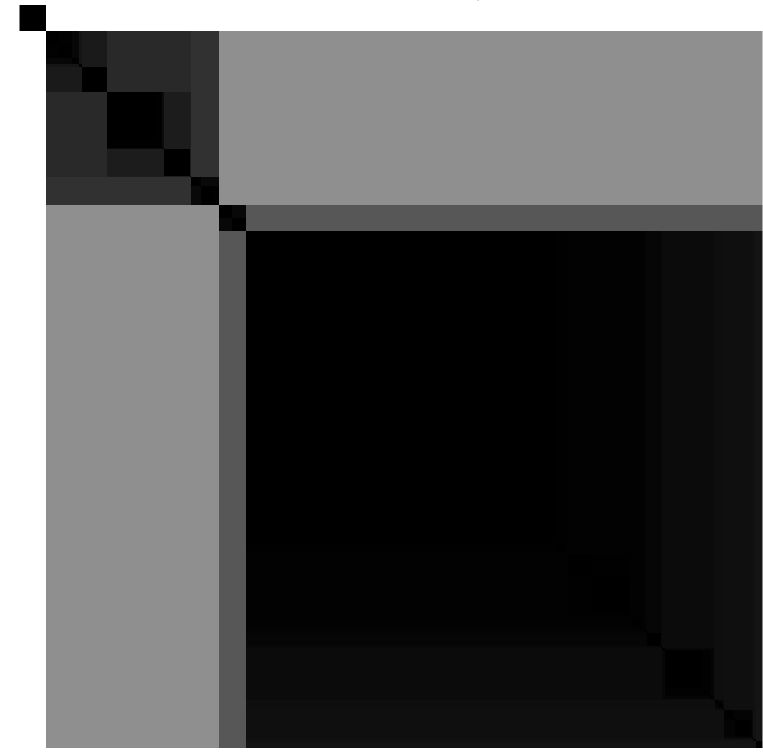}}
\caption{Soybean dataset image ; (a) iVAT image (PA: 27.6\%); (b) ConiVAT (PA: 46.3\%).}%
\label{Fig:Soybean_Dataset}
\end{figure}

Our last example in this experiment is on the SOYBEAN (large version) dataset which has $562$ instances with $35$ nominal attributes that are labeled in $19$ classes. The corresponding iVAT image in~\Fig~\ref{Fig:Soybean_Dataset} shows one big, $7-8$ small size, and a few tiny and singleton dark blocks suggesting more than $10$ clusters in the dataset. In contrast, the ConiVAT suggests $3$ dark block on low-resolution and $5$ dark blocks on high-resolution suggesting $3-5$ clusters in the dataset. Note that, for the SOYBEAN dataset, although the iVAT estimates the presumed number of clusters better than the ConiVAT, the ConiVAT achieves higher clustering accuracy ($46.3\%$ compared to iVAT ($27.6\%$). This is probably because SL-aligned partitions in iVAT image are ruined by several singleton clusters.

To summarize, ConiVAT can enhance the quality of iVAT images using a few randomly generated constraints and can reveal the hidden cluster structure, especially for datasets having noisy bridge between clusters. Thus, it provides much reliable estimate of the potential number of clusters present in the data, compared to iVAT, and consequently, it improves the SL clustering accuracy.

\subsubsection{Comparison of different constraint-based clustering methods}
In this experiment, we compare ConiVAT with two standard \textit{hierarchical agglomerative clustering} (HAC) algorithms using (i) SL supervised hierarchical clustering algorithm, (iii) \textit{Ultrametric tree using transitive dissimilarity}  (UltraTran)~\cite{zheng2011semi} ; (iv) \textit{Constraint CL} (CCL)~\cite{klein2002instance}; and (v) \textit{semi-supervised SL} (SSL)~\cite{reddy2016semi}. Table~\ref{table:Comparison} shows the comparison of these six clustering algorithms based on the PA (\%). We observe that: (i) For all the presented datasets except SYNTH1, the ConiVAT significantly outperforms the other five clustering approaches based on the partition accuracy. Specifically, for SYNTH2, VOTING, MUSH, and PIMA, ConiVAT beats other semi-supervised approaches by a significant margin in clustering accuracy; (ii) The CCL method achieves the highest clustering accuracy (93.9\%) for SYNTH1; (iii) The performance improvement from ConiVAT is not very significant on SOYBEAN dataset. This is probably because the constraints generated from the SOYBEAN dataset might not have representation from many of the $19$ clusters.

\begin{figure}
\captionsetup[subfigure]{justification=centering}
\centering
\includegraphics[width=0.53\textwidth]{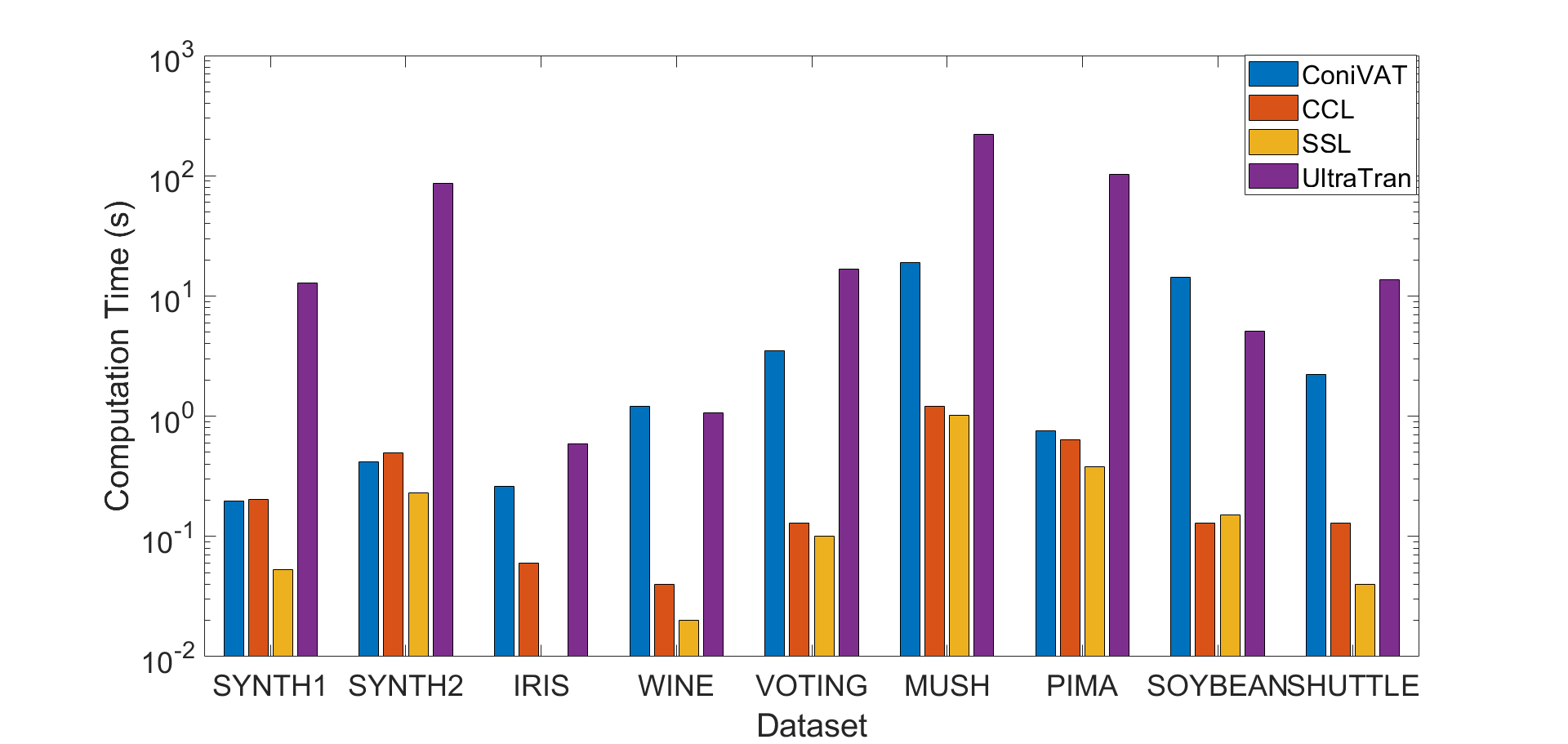}
\caption{CPU-time (log-scale on $y$-axis) of four semi-supervised algorithms.}
\label{Fig:Time_performance}
\end{figure}

Fig~\ref{Fig:Time_performance} shows the clustering performance of all four semi-supervised algorithms based on the run-time (in seconds). Among them, UltraTran has the longest and SSL has the shortest execution time. Overall, the ConiVAT has the second-highest execution-time. In our experiment, we observed that the run-time of ConiVAT is comparable to the run-time of CCL and SSL algorithms for low-dimensional datasets; however, for relatively high-dimensional data such as WINE, VOTING, SOYBEAN and SHUTTLE, the metric-learning step in ConiVAT takes longer to terminate and learn a $p \times p$ full weight matrix, thus increasing the overall execution-time of ConiVAT. Note that, although ConiVAT runs a bit slower than CCL and SSL, it outperforms them by a good margin in clustering accuracy. In summary, integrating the constraints in iVAT (ConiVAT framework) significantly outperforms the other semi-supervised approaches in terms of clustering accuracy. Balanced against this improvement is the time cost of ConiVAT, especially for high-dimensional datasets. Moreover, unlike the other three approaches, ConiVAT provides visual evidence about potential cluster structure in the dataset.

\subsubsection{Benefits from each step of ConiVAT}
\begin{figure}
\captionsetup[subfigure]{justification=centering}
\centering
 \includegraphics[width=0.5\textwidth]{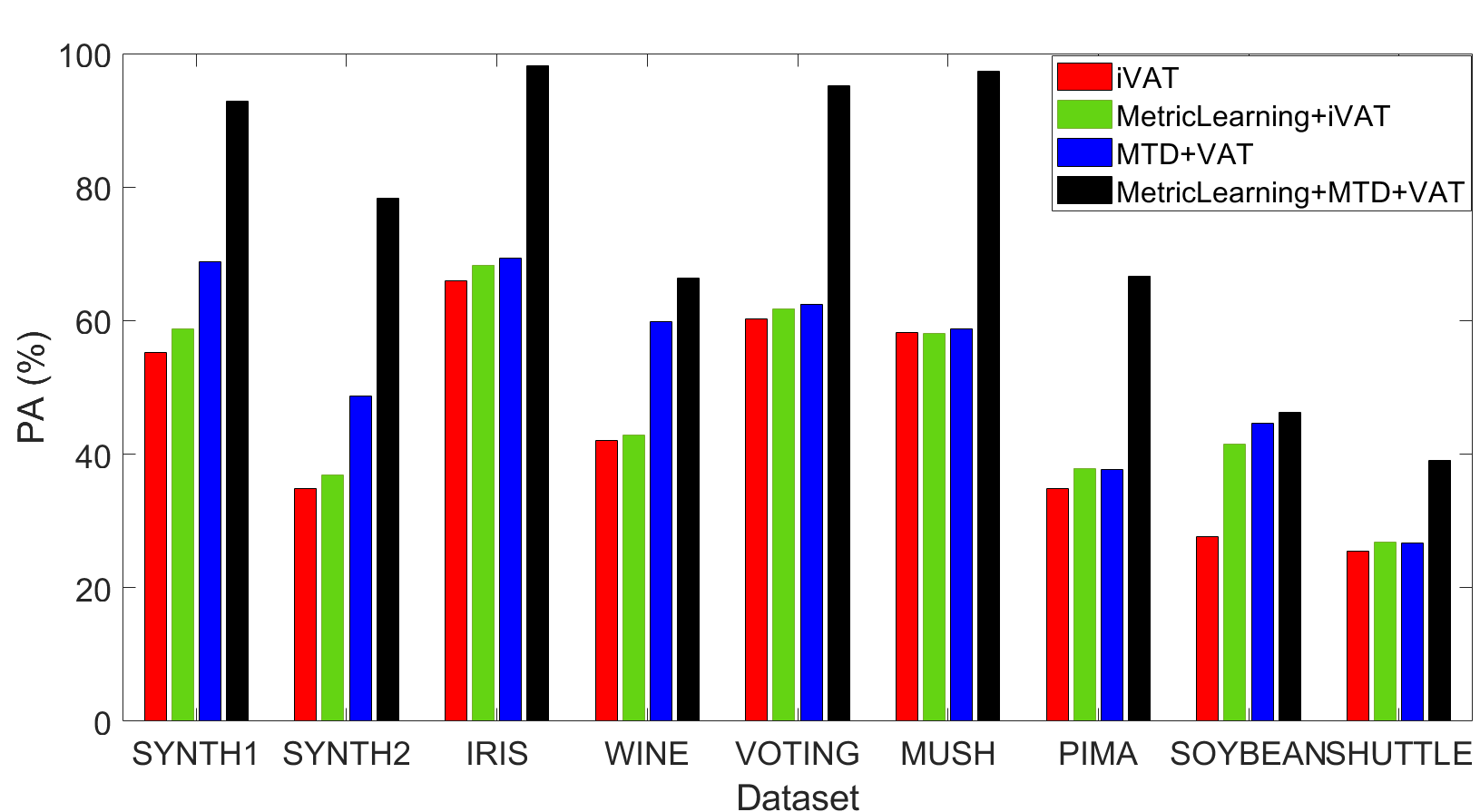}
\caption{Performances of various ablated instances of ConiVAT and their combination (ConiVAT).}
\label{Fig:AblationStudy}
\end{figure}

In this experiment, we evaluated the utility of individual steps in the ConiVAT algorithm. In this regard, we conducted this experiment using the various ablated instances of ConiVAT and its unified approach viz., (i) iVAT~\cite{havens2012efficient} (no constraints); (ii) metric-learning + iVAT; (iii) MTD + VAT (path-based transform of $D$ (with must link pair entries as $0$s) followed by VAT) ; (iv) metric learning + MTD + VAT i.e. ConiVAT. These four methods are compared by their relative clustering  accuracy.~\Fig~\ref{Fig:AblationStudy} shows the clustering accuracy for each of the above four models for all nine datasets. The MTD step provides a substantial improvement over iVAT. Moreover, although both metric-learning and MTD individually improve clustering performance, the unified approach (all the steps combined) significantly outperforms any of its components for all the datasets. 

\subsubsection{Effect of constraints on ConiVAT}
\begin{figure}
\captionsetup[subfigure]{justification=centering}
\centering
\subfloat[]{\includegraphics[width=0.5\textwidth]{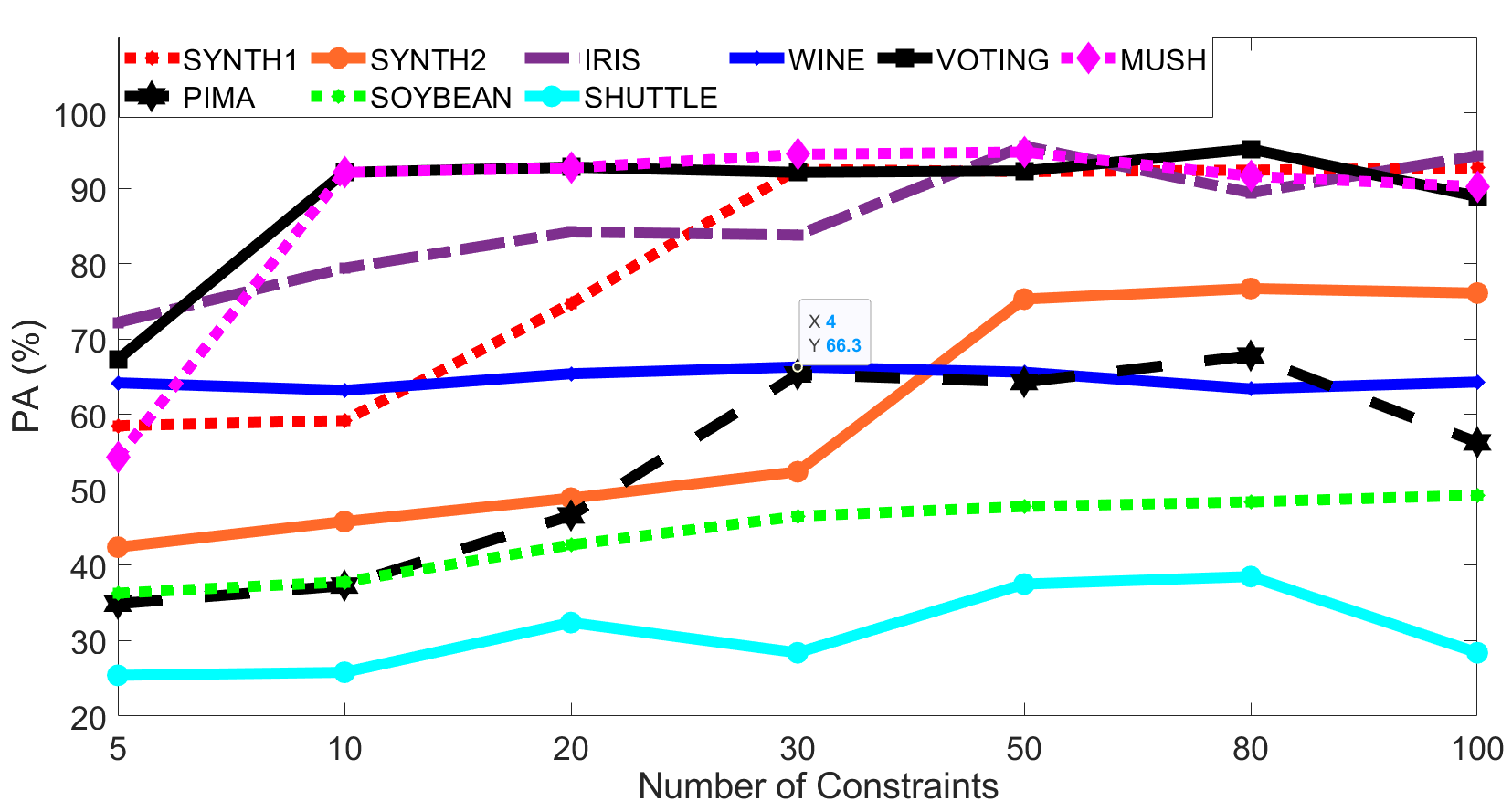}}\\
\subfloat[]{\includegraphics[width=0.5\textwidth]{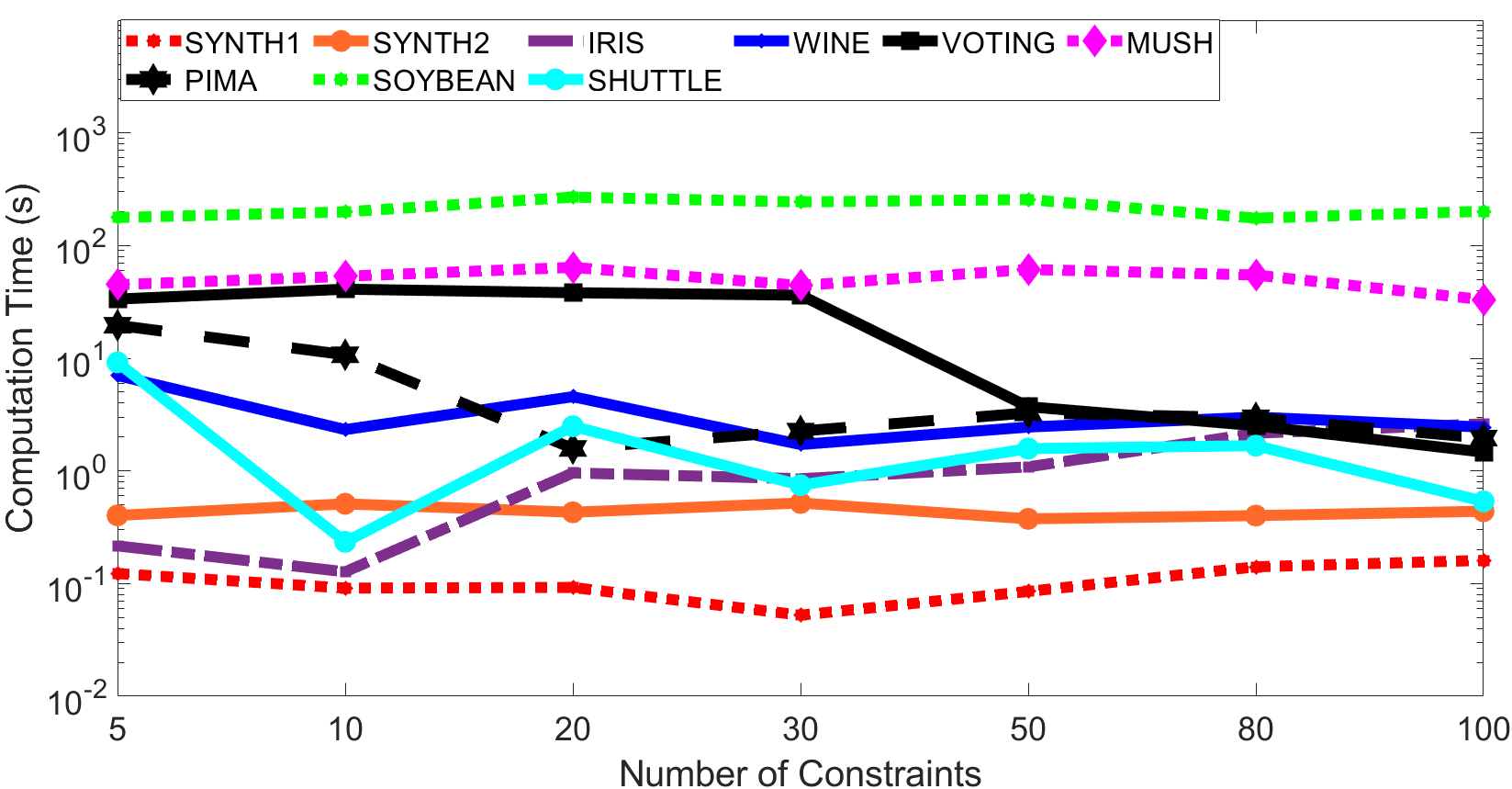}}
\caption{Effect of a varying number of constraints on ConiVAT (a) clustering accuracy ; and (b) computation time.}%
\label{Fig:VaryingConstraints}
\end{figure}

In this experiment, we investigate the effect of the number of constraints on ConiVAT.~\Fig~\ref{Fig:VaryingConstraints} shows the clustering accuracy and computation time of ConiVAT for a varying number of constraints i.e. $5, 10, 20, 30, 50, 80$, and $100$, for all nine datasets. We can see that, even with a minimal number of randomly generated constraints, ConiVAT achieves  a significant boost in clustering accuracy. The computation time of coniVAT remains almost same as the number of constraint increases. 
 Although the clustering accuracy of ConiVAT increases with the number of constraints, it does not monotonically increase. This is probably because, like other semi-supervised approaches, the performance of ConiVAT also depends on the quality of constraints. Clearly, not all the constraints will be useful, particularly when the corresponding  relationships can automatically and easily be deduced by a clustering algorithm. Only a few constraints, which greatly assist the algorithm to identify complex and difficult patterns, will be more useful.

\section{Future Work}\label{sec:futurework}
Our experiments suggest that even randomly generated constraints can improve the performance of ConiVAT. From this it follows that intelligently selected constraints using active learning~\cite{basu2004active,grira2008active} e.g. \textit{farthest first traversal} (maximin sampling)~\cite{bilenko2004integrating}, could be even more beneficial. We intend to explore active learning to generate useful constraints in our framework. Also, there are some methods~\cite{masud2019generate} which can generate useful pairwise constraints from the unlabeled data. We intend to examine the effect of such constraints in our future work. 

A recent review article~\cite{gera2015parameterized} on semi-supervised clustering emphasized the need of  semi-supervised clustering algorithms for voluminous data. Like existing semi-supervised clustering algorithms,  ConiVAT may take a significant time for big data. We observed in our experiments that ConiVAT takes a relatively long time for high-dimensional datasets. Therefore, in our future work, we aim to develop an efficient and scalable version of ConiVAT to handle large volumes of high-dimensional datasets. Since \textit{scalable iVAT} (siVAT) and its variants been proposed in~\cite{hathaway2006scalable,rathore2018rapid,rathore2018approximate,rathore2020visual} as scalable versions of iVAT for big data, we aim to integrate them with ConiVAT framework in our future work, to handle large volumes of high-dimensional data. 

\section{Conclusions}\label{sec:conclusion}
This article proposed a semi-supervised version of the iVAT algorithm, called ConiVAT, that integrates  metric-learning and constraints with iVAT in a principled manner. The ConiVAT model learns an underlying similarity metric from the constraints such that it satisfies the input constraints. We performed extensive experiments on nine datasets, including seven real-datasets. We demonstrated that the ConiVAT enhances the quality of iVAT images and can reveal the hidden structure, especially for complex datasets and datasets with "noisy bridges" between clusters, better than iVAT does. Moreover, ConiVAT also improves the SL clustering accuracy by using only a few numbers of randomly generated constraints as background knowledge.

We compared SL clustering based on the ConiVAT MST with three other hierarchical semi-supervised clustering approaches and two standard hierarchical clustering algorithms on nine datasets. Experimental results suggest that SL using ConiVAT MST significantly outperforms the other five clustering approaches in terms of clustering accuracy. 
\bibliographystyle{IEEEtran}
\bibliography{Bibliography}
\vskip -2\baselineskip plus -1fil

\begin{IEEEbiography}[{\includegraphics[width=1in,height=1.25in,clip,keepaspectratio]{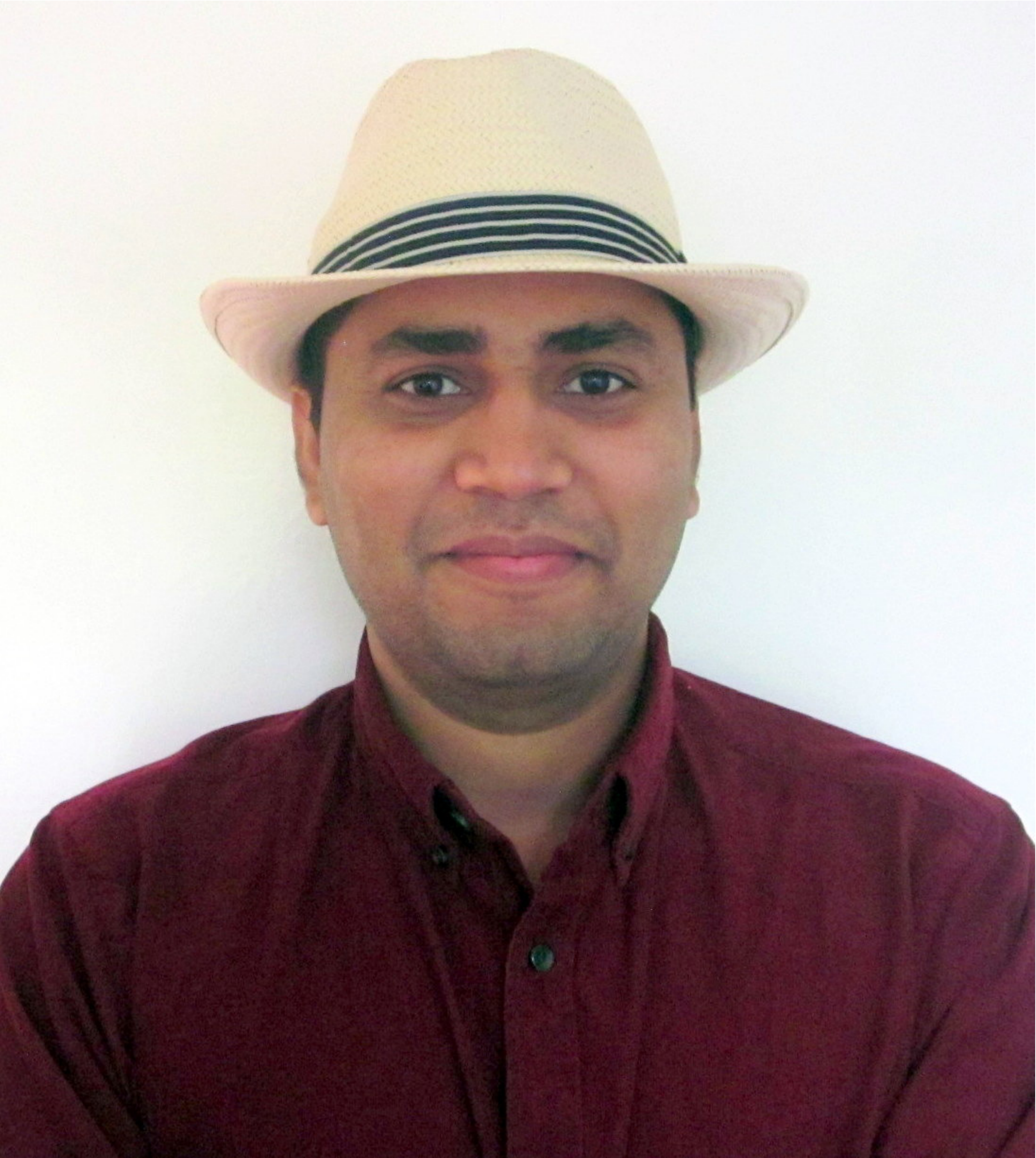}}]%
{Punit Rathore} is a Postdoctoral Research Fellow in Senseable City Lab, Department of Urban Studies and Planning at Massachusetts Institute of Technology (MIT) Cambridge, Previously, he worked as a Postdoctoral Researcher at Institute of Data Science, National University of Singapore (NUS) Singapore. He received the M.Tech degree in Instrumentation Engineering from Indian Institute of Technology, Kharagpur, India in 2011, and Ph.D degree from Department of Electrical and Electronics Engineering, the University of Melbourne, Melbourne, Australia in 2019. His research interests include big data cluster analysis, anomaly detection, urban data analytics, and Internet of Things.
\end{IEEEbiography}
\begin{IEEEbiography}[{\includegraphics[width=1in,height=1.2in,clip,keepaspectratio]{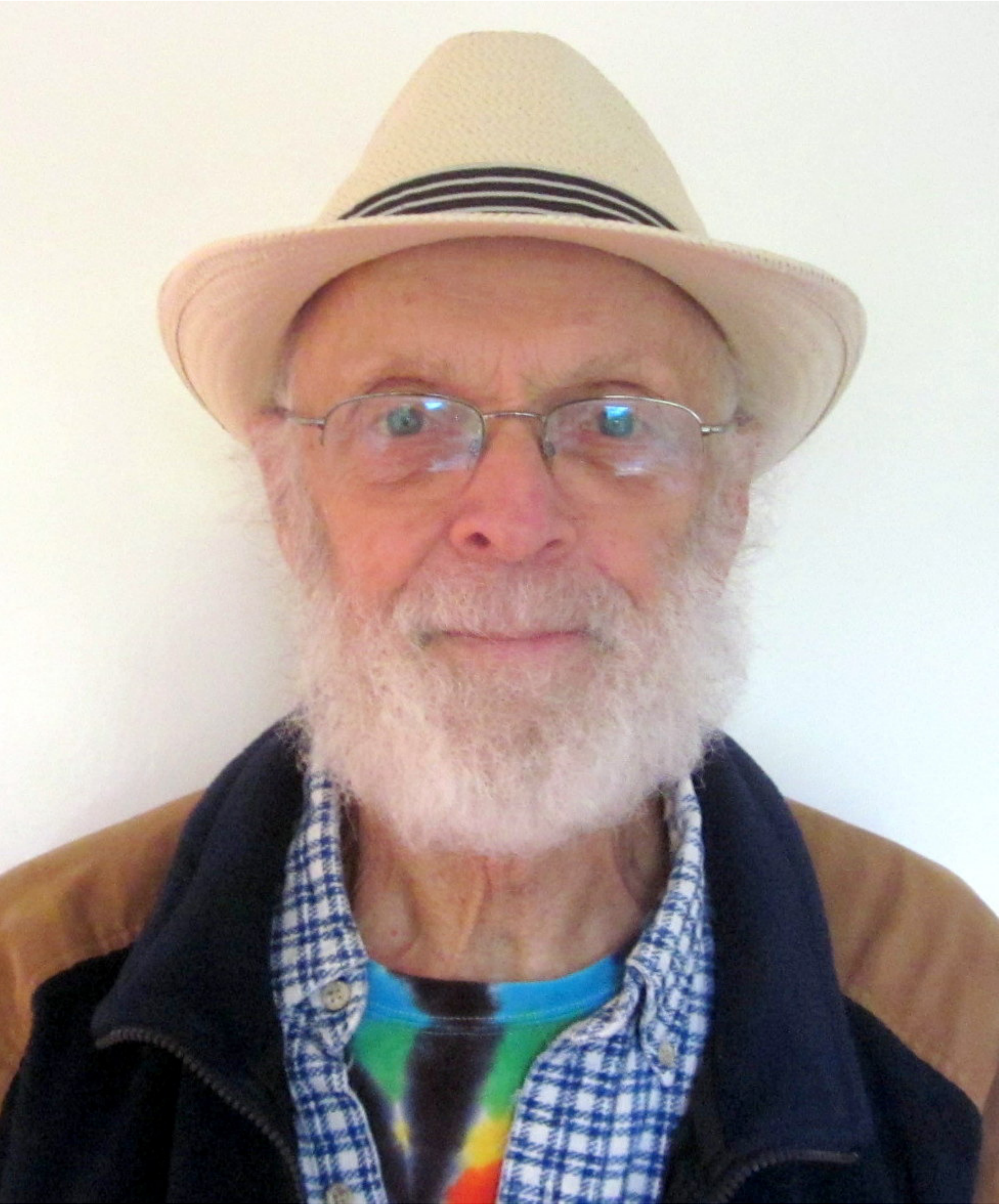}}]%
{James C. Bezdek (LF'10)} received the PhD in Applied Math, Cornell University, $1973$. Jim is past President of NAFIPS (North
American Fuzzy Information Processing Society), IFSA (International Fuzzy Systems Association) and the IEEE CIS (Computational Intelligence Society as the NNC): founding editor the International Journal of Approximate Reasoning and the IEEE Transactions on Fuzzy Systems: Life fellow of the IEEE and IFSA; recipient of the IEEE 3rd Millennium, IEEE CIS Fuzzy Systems Pioneer, IEEE Frank Rosenblatt TFA and the Kempe de Feret IPMU awards. He retired in 2007. His research interests include optimization, pattern recognition, and big data clustering.
\end{IEEEbiography}
\vskip -2\baselineskip plus -1fil
\begin{IEEEbiography}[{\includegraphics[width=1in,height=1.2in,clip,keepaspectratio]{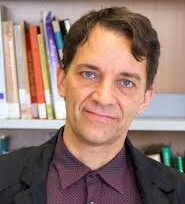}}]%
{Paolo Santi} is a Principal Research Scientist at MIT Senseable City Lab where he leads the MIT/Fraunhofer Ambient Mobility initiative, and a Senior Research
at the Istituto di Informatica e Telematica, CNR, Pisa. Dr. Santi holds a "Laurea" degree and Ph.D. in computer Science from the University of Pisa, Italy. Dr. Santi is a member of the IEEE Computer Society and has recently
been recognized as Distinguished Scientist by the Association for Computing Machinery. His research interest is in the modeling and analysis of complex
systems ranging from wireless multi-hop networks to sensor and vehicular networks and, more recently, smart mobility and intelligent transportation systems. In these fields, he has contributed more than 120 scientific papers and two books. Dr. Santi has been involved in the technical and organizing committee of several conferences in the field, and he is/has been an Associate Editor of the IEEE Transactions on Mobile Computing, the IEEE Transactions on Parallel and Distributed Systems, and Computer Networks.
\end{IEEEbiography}
\vskip -2\baselineskip plus -1fil
\begin{IEEEbiography}[{\includegraphics[width=1in,height=1.2in,clip,keepaspectratio]{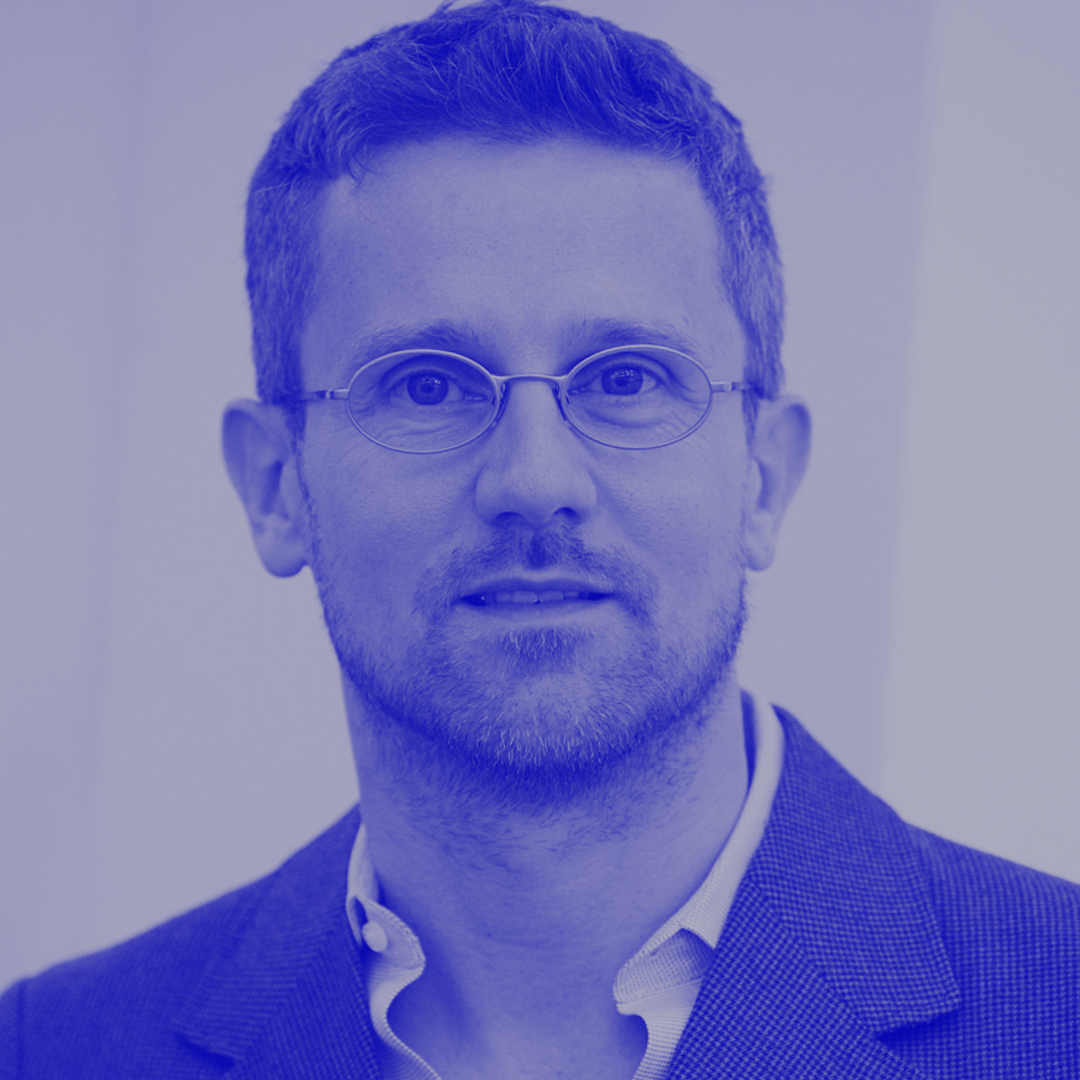}}]%
{Carlo Ratti} received the Master's degree in Civil
Engineering from the Politecnico di Torino and the
Ecole Nationale des Ponts et Chaussees, Paris, and
the M.Phil. and Ph.D. degrees from the University of
Cambridge, U.K. As an Architect and an Engineer
by training, he teaches at the Massachusetts Institute
of Technology, where he directs the Senseable City
Laboratory and is a Founding Partner of the international design office Carlo Ratti Associati. His work
has been exhibited worldwide at venues such as the
Venice Biennale; the Design Museum, Barcelona;
the Science Museum, London; and The Museum of Modern Art, New York City. He has co-authored over 500 publications and holds several patents. He is currently serving as both a member for the World Economic Forum Global Future Council on Cities and a Special Adviser on Urban Innovation for the European Commission.
\end{IEEEbiography}

\end{document}